%% file: 0-main.tex
\documentclass[conference]{IEEEtran}
\IEEEoverridecommandlockouts
\usepackage{cite}
\usepackage{amsmath,amssymb,amsfonts}
\usepackage{algorithmic}
\usepackage{graphicx}
\usepackage{textcomp}
\usepackage[table]{xcolor}
\usepackage{xcolor}

\usepackage{algorithm}
\usepackage{tabularx}
\usepackage{multirow}
\usepackage{subfigure}
\usepackage{caption}
\usepackage{bm}
\usepackage{color}
\usepackage{afterpage}
\newcommand\scalemath[2]{\scalebox{#1}{\mbox{\ensuremath{\displaystyle #2}}}}
\usepackage{booktabs}
\usepackage{bbding}
\usepackage{makecell}
\usepackage{amsthm}
\usepackage[T1]{fontenc}
\usepackage{rotating}
\usepackage{float}
\usepackage{soul}
\usepackage[flushleft]{threeparttable}

\theoremstyle{definition}
\newtheorem{proposition}{Proposition}

\theoremstyle{definition}

\def\BibTeX{{\rm B\kern-.05em{\sc i\kern-.025em b}\kern-.08em
    T\kern-.1667em\lower.7ex\hbox{E}\kern-.125emX}}
\begin{document}

\title{Graph Soft-Contrastive Learning via Neighborhood Ranking}

\author{\IEEEauthorblockN{Zhiyuan Ning$^{1,2,\dag}$, Pengfei Wang$^{1,2,\dag}$, Pengyang Wang$^{3,*}$, Ziyue Qiao$^{6}$, Wei Fan$^{4}$, Denghui Zhang$^{5}$,\\ Yi Du$^{1,2}$, Yuanchun Zhou$^{1,2}$}
\IEEEauthorblockA{$^1$Computer Network Information Center, Chinese Academy of Sciences, $^2$University of Chinese Academy of Sciences \\
$^3$IOTSC, University of Macau, $^4$University of Central Florida, $^5$Rutgers University\\
$^6$The Hong Kong University of Science and Technology (Guangzhou)\\
\{ningzhiyuan, pfwang\}@cnic.cn, pywang@um.edu.mo, ziyuejoe@gmail.com\\ weifan@knights.ucf.edu, dhzhangai@gmail.com, \{duyi, zyc\}@cnic.cn
\thanks{$^{\dag}$Equal contribution. $^{*}$Corresponding author. This work was done when Zhiyuan Ning did the internship at IOTSC, University of Macau.
}
}
}


\maketitle

\input{1-abstract.tex}


\input{2-intro.tex}
\input{3-preliminary.tex}

\input{4-method.tex}

\input{5-experiment.tex}

\input{6-related-work.tex}
\input{7-conclusion.tex}

\appendix
\input{8-appendix.tex}


\newpage
\bibliographystyle{IEEEtran}
\bibliography{9-ref}
\end{document}

%% file: 1-abstract.tex
\begin{abstract}
Graph Contrastive Learning (GCL) has emerged as a promising approach in the realm of graph self-supervised learning. 
Prevailing GCL methods mainly derive from the principles of contrastive learning in the field of computer vision: modeling invariance by specifying absolutely similar pairs. 
However, when applied to graph data, this paradigm encounters two significant limitations: 
(1) the validity of the generated views cannot be guaranteed: graph perturbation may produce invalid views against semantics and intrinsic topology of graph data; 
(2) specifying absolutely similar pairs in the graph views is unreliable: for abstract and non-Euclidean graph data, it is difficult for humans to decide the absolute similarity and dissimilarity intuitively. 
Despite the notable performance of current GCL methods, these challenges necessitate a reevaluation: Could GCL be more effectively tailored to the intrinsic properties of graphs, rather than merely adopting principles from computer vision? 
In response to this query, we propose a novel paradigm, Graph Soft-Contrastive Learning (GSCL). 
This approach facilitates GCL via neighborhood ranking, avoiding the need to specify absolutely similar pairs. 
GSCL leverages the underlying graph characteristic of diminishing label consistency, asserting that nodes that are closer in the graph are overall more similar than far-distant nodes. 
Within the GSCL framework, we introduce pairwise and listwise gated ranking InfoNCE loss functions to effectively preserve the relative similarity ranking within neighborhoods. 
Moreover, as the neighborhood size exponentially expands with more hops considered, we propose neighborhood sampling strategies to improve learning efficiency. 
Our extensive empirical results across 11 commonly used graph datasets—including 8 homophily graphs and 3 heterophily graphs—demonstrate GSCL's superior performance compared to 20 state-of-the-art GCL methods. 
\end{abstract}

%% file: 2-intro.tex
\section{Introduction}
\begin{figure}[!t]
\centering
\includegraphics[width=1.0\linewidth]{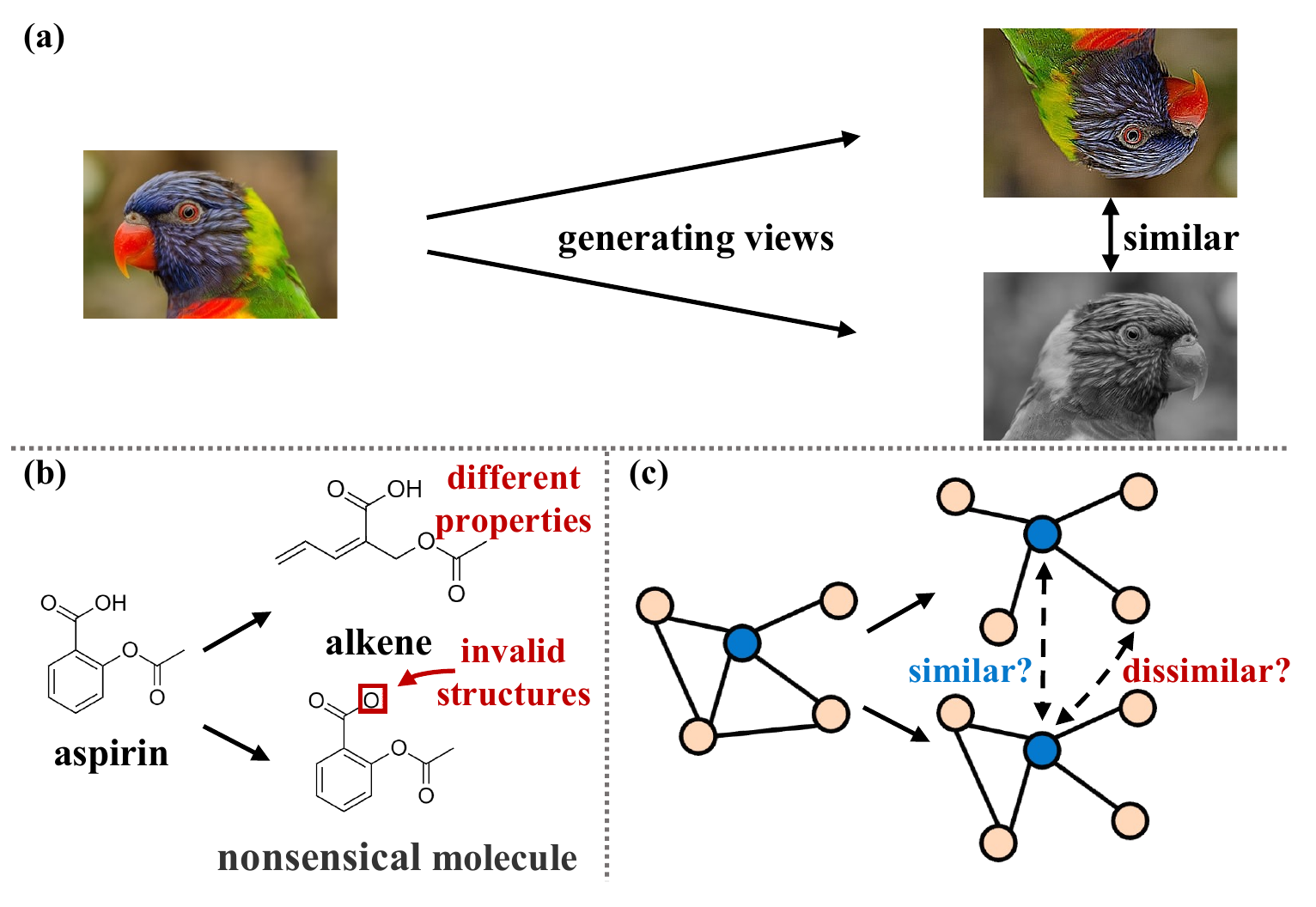}
\caption{A comparison of CL for image and graph data. 
Figure (a) indicates that two augmented views of the same parrot image keep semantic similarity. 
Figure (b) shows the view generation in molecular graphs, which yields distinct properties and invalid structures. 
Figure (c) shows that it is difficult for humans to judge the similarity of generated views for abstract and non-euclidean graph data.
}
\label{fig:example}
\vspace{-4.0mm}
\end{figure}

The concept of Contrastive Learning (CL) has its roots in the field of computer vision and aims to learn such an embedding space in which similar samples are kept close while dissimilar ones are far apart~\cite{hjelm2018learning,chen2020simple,he2020momentum}. 
To implement CL in an unsupervised setting, two augmented views of the same data example are generated (as shown in Figure~\ref{fig:example}) and the similarity between the encoded representations of the two views will increase. 
Based on such a view generation framework, CL aims to model \textit{\textbf{invariance}}, {\it i.e.}, two observations of the same concept should produce the same outputs~\cite{chen2021exploring}. 
As illustrated in Figure~\ref{fig:example} (a), for an image of a parrot, two views are generated through data augmentation. 
With the aid of human visual perception, it is apparent that these two views both depict the concept of a parrot, and thus, it is logical to pull them closer in embedding space.

Graph Contrastive Learning (GCL)~\cite{Velickovic2019DeepGI,Zhu2020DeepGC, thakoor2021large} adopts the similar paradigm in the computer vision domain~\cite{chen2020simple,he2020momentum,chen2021exploring, grill2020bootstrap}: utilizing graph augmentation techniques~\cite{Zhu2021AnES} to generate graph views (e.g., node dropping~\cite{you2020graph}, edge removing~\cite{Zhu2020DeepGC}, diffusion~\cite{hassani2020contrastive}, etc.), and subsequently optimizes the CL loss function commonly used in computer vision~\cite{hjelm2018learning,chen2020simple,grill2020bootstrap} to model invariance. 
Although traditional GCL methods achieve impressive performance, how to generate label-invariant and optimal views for a graph through data augmentation is non-trivial~\cite{Xia2022SimGRACEAS,Hou2022GraphMAESM,Lee2022AugmentationFreeSL}. 
Follow-up efforts have been made to enable augmentation-free GCL in several ways to avoid generating views through data augmentation~\cite{Zhang2022COSTACF,Xia2022SimGRACEAS,Yu2022AreGA,Mo2022SimpleUG,Lee2022AugmentationFreeSL}. 
Despite variations in motivations and implementations, the fundamental ideas underlying all these GCL methods are the same, \textit{\textbf{they all aim to model invariance in an absolute manner, {\it i.e.}, specifying absolutely similar pairs in the graph views. }}

However, different from image data, graph data is discrete, abstract and non-Euclidean~\cite{thakoor2021large}. 
Modeling invariance for graph by specifying absolutely similar pairs in the graph views is less appropriate due to the following reasons: 
\textbf{(1)} {\it \ul{The validity of the generated views cannot be guaranteed}}: 
regardless of the strategies implemented for generating graph views, maintaining the conceptual validity of these views presents a considerable challenge. 
Primarily, two issues are encountered. 
First, these views could potentially exhibit characteristics that markedly diverge from the properties inherent in the original graph~\cite{Lee2022AugmentationFreeSL}. 
Second, there exists the risk of generating views that are nonsensical—that is, they lack any conceivable real-world evidence. 
For example, when examining molecular graphs as represented in Figure~\ref{fig:example} (b), the deletion of nodes or edges from the aspirin molecule could lead to the creation of an alkene with disparate functionalities, or, more detrimentally, yield a structurally invalid, nonsensical molecule. 
\textbf{(2)} {\it \ul{Specifying absolutely similar pairs in the graph views is unreliable}}: 
although we humans can use domain knowledge to understand and analyze some graphs, such as molecular graphs. 
But for most graphs which are only abstract relationships between objects, our understanding of them is not as intuitive as our perception of images~\cite{Hou2022GraphMAESM}. 
We generally lack the capacity to make direct judgments about the absolute similarity between nodes or graphs, as depicted in Figure~\ref{fig:example} (c). 
Consequently, it is unreliable to assume that two generated views of the same node or graph should inherently bear similarity and thereby occupy closer proximity within the embedding space. 
While existing GCL methods exhibit commendable performance, the aforementioned analysis prompts us to consider: \textit{\textbf{whether GCL could be designed more rationally by considering the unique characteristics of graphs themselves rather than directly adapting the CL ideas from computer vision?}}

\begin{figure}[!t]
\centering
\includegraphics[width=1.0\linewidth]{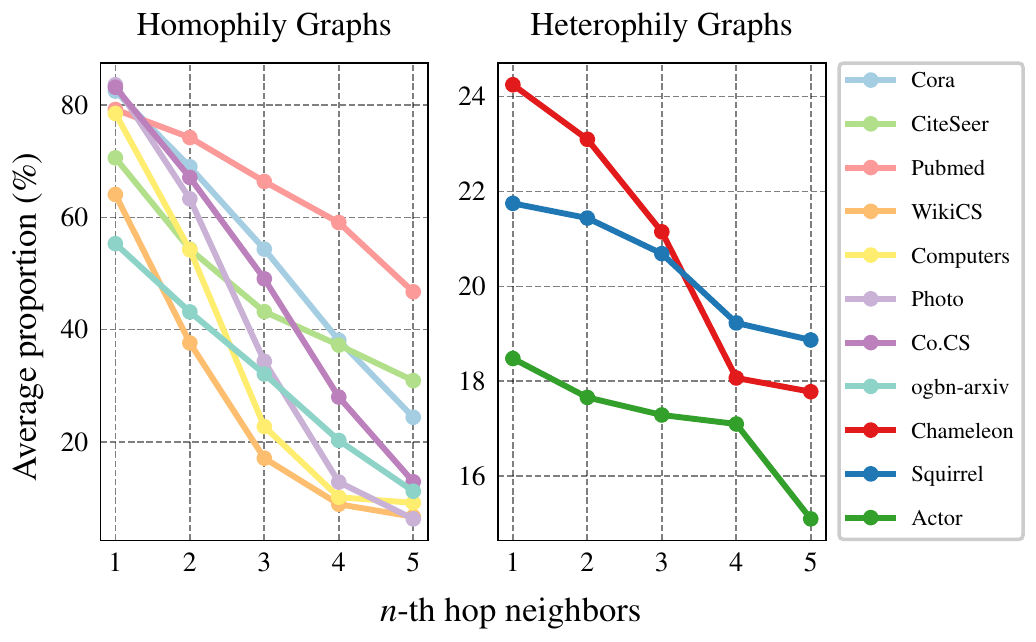}
\caption{
The average proportion of neighbors sharing the same label as the anchor node (the  label consistency) in each hop.
}
\label{fig:motivation}
\vspace{-4.0mm}
\end{figure}

Graph data inherently holds a wealth of valuable structural properties, one of the most significant being homophily~\cite{mcpherson2001birds, bramoulle2012homophily}, which implies that similar nodes are often more likely to be interconnected. 
This phenomenon is prevalent across various types of real-world graphs, including but not limited to friendship, political, and citation networks~\cite{mcpherson2001birds, gerber2013political, ciotti2016homophily}, which are collectively referred to as homophily graphs. 
Conversely, many graphs exhibit clear  heterophily, where connections mainly exist between dissimilar nodes~\cite{Pei2020GeomGCNGG,zhu2020beyond}. 
Despite the apparent differences in homophily between these two types of graphs, a shift in focus to a broader neighborhood range reveals a surprising uniformity. 
Specifically, Figure~\ref{fig:motivation} shows the average proportion of neighbors sharing the same label as the anchor node in each hop from 11 commonly used public graph datasets (including 8 homophily graphs and 3 heterophily graphs). 
The results statistically indicate the same trend in all graphs: closer neighbors have a greater probability of having the same label as the anchor node, and the probability decreases as the distance gets farther. 
We call such trend as {\it \textbf{the overall diminishing trend of label consistency}}, which provides us with a convincing perspective to determine the similarity, {\it i.e.}, nodes that are closer in the graph are overall more similar than far-distant nodes. 
Motivated by such \textbf{\textit{relative}} decisions of similarity, we present a novel CL paradigm for graph data, namely Graph Soft-Contrastive Learning (GSCL), that conducts CL without specifying any absolutely similar pairs. 
Instead, GSCL maintains the similarity ranking relationship in the neighborhood that the less distance/hops to the anchor node, the more similar the neighbors are to the anchor node overall. 
To achieve this, we propose pairwise and listwise gated ranking InfoNCE loss functions that robustly maintain the neighborhood ranking relationship. 
As more hops are considered, the neighborhood size would expand exponentially. 
Therefore, to further improve the learning efficiency, we propose neighborhood sampling strategies to reduce the computational burden.

In summary, our contributions can be listed as follows:
\begin{itemize}
\item We analyze and summarize that current GCL methods all model invariance in an absolute manner, but such idea is less natural and inappropriate for graph-structured data.
\item We propose GSCL, a soft contrastive learning paradigm for graph data, that achieves our goal of utilizing the inherent properties of graph to design GCL, by preserving the neighborhood ranking relationship, which reflects the overall diminishing trend of label consistency in both homophily graphs and heterophily graphs. 
\item To implement GSCL, we develop two novel loss functions, the pairwise and listwise gated ranking InfoNCE loss as the learning objective. Besides, we propose two neighborhood sampling strategies to further accelerate the learning process of GSCL.
\item To verify the rationality of GSCL, we conduct extensive experiments on 11 real-world datasets (including 8 homophily graph datasets and 3 heterophily graph datasets) against 20 baselines. The results demonstrate GSCL, without specifying any absolutely similar pairs, can achieve superior performance compared with current GCL methods and obtain state-of-the-art in all datasets. 
\end{itemize}

%% file: 3-preliminary.tex
\section{Preliminary}

\subsection{Notation}
Formally, let $\mathcal{G}=\{\mathcal{V}, \mathcal{E}, \mathbf{X}, \mathbf{A}\}$ denotes a graph with $N$ nodes, where $\mathcal{V}=\{v_i\}|^{N}_{i=1}$ denotes the node set, $\mathcal{E} \subseteq \mathcal{V} \times \mathcal{V}$ denotes the edge set, $\mathbf{X}\in \mathbb{R}^{N \times D}$ denotes the $D$ dimensional node feature matrix, and $\mathbf{A} \in \mathbb{R}^{N \times N}$ denotes the adjacency matrix where $\mathbf{A}_{ij}=1$ iff $(v_i, v_j) \in \mathcal{E}$ and $\mathbf{A}_{ij}=0$ otherwise. 
For each node $v_i$ in the graph $\mathcal{G}$, $\mathcal{N}(v_i)^{[n]}$ denotes the set of all its $n$-th hop neighbors. 
And when the neighborhood~\footnote{In this paper, neighbor refers to a node that is close to the anchor node, while neighborhood refers to the area around the location of the anchor node.} range of $v_i$ we consider is $k$, all its neighbors can be organized from small to large in hops as $\mathbb{N}(v_i)^{[k]}=\{ \mathcal{N}(v_i)^{[1]}, \mathcal{N}(v_i)^{[2]}, \cdots, \mathcal{N}(v_i)^{[k]}, \mathcal{N}(v_i)^{[k+1]}\}$, where $\mathcal{N}(v_i)^{[k+1]}$ denotes all the nodes with a distance larger than $k$ hops from $v_i$. 

In this paper, we will study the problem of learning node representations in a unsupervised setting via GCL. 
Our objective is  to unsupervisedly learn an optimal graph encoder $f: (\mathbf{X}, \mathbf{A}) \rightarrow \mathbf{H}\in \mathbb{R}^{N \times D^{'}}$, that can transform the original node feature $\mathbf{X}$ to lower dimensional ($D \gg D^{'}$) hidden representations $\mathbf{H}$, which can benefit the downstream supervised or semi-supervised tasks, e.g., node classification.

\subsection{Graph Homophily}\label{sec:homo}
Most common graphs follow the principle of homophily, {\it i.e.}, ``birds of a feather flock together"~\cite{mcpherson2001birds}. 
Specifically, graph homophily refers to the theory in network science which states that similar nodes may be more likely to attach to each other than dissimilar ones~\cite{mcpherson2001birds, bramoulle2012homophily}, which suggests that connected nodes often belong to the same class. 
This phenomenon is ubiquitous in a variety of real-world networks, including but not limited to friendship networks~\cite{mcpherson2001birds}, political networks~\cite{gerber2013political}, citation networks~\cite{ciotti2016homophily}, and others. 
In addition, graph neural network (GNN), a prominent approach for learning representations for graph structured data nowadays, is believed to implicitly assume strong homophily~\cite{zhu2021graph, zhu2020beyond}. 
Homophily is often calculated as the the average proportion of edge-label consistency of all nodes~\cite{Pei2020GeomGCNGG,wang2022augmentation}. 
Formally, the homophiliy metric ($\mathit{HM}$) of a graph can be defined as follows: 
\begin{equation}
\mathit{HM}=\frac{1}{N} \sum_{v_i \in \mathcal{V}} \frac{\left|\left\{v_j:\left(v_i, v_j\right) \in \mathcal{E} \wedge y_i=y_j\right\}\right|}{\left|\left\{v_j:\left(v_i, v_j\right) \in \mathcal{E}\right\}\right|}.
\end{equation}
The values of $\mathit{HM}$ are in the range $[0, 1]$, values close to 1 indicate high homophily and values close to 0 indicate low homophily. 
Note that while many graphs in the real world have high homophily, there are also some kinds of graphs which have low homophily~\cite{Pei2020GeomGCNGG,zhu2020beyond}. 
By convention, we refer to graphs with high homophily as homophily graphs and graphs with low homophily as heterophily graphs. 
We provide the type and $\mathit{HM}$ of graphs considered in this paper in Table~\ref{tab:datasets_statistics}. 
We can observe that all homophily graphs' $\mathit{HM}$s are greater than 0.55, yet none of heterophily graphs' $\mathit{HM}$s exceed 0.25. 

\subsection{The Overall Diminishing Trend of Label Consistency}\label{sec:lc}
$\mathit{HM}$ is essentially calculating the average proportion of first hop neighbors that share the same label as the anchor node. 
Now, we extend such proportion from the first hop neighbors to any $n$-th hop neighbors and define it as the label consistency ($\mathit{LC}$) in the $n$-th hop: 
\begin{equation}
\mathit{LC}\left(n\right)=\frac{1}{N} \sum_{v_i \in \mathcal{V}} \frac{\left|\left\{v_j: v_j \in \mathcal{N}(v_i)^{[n]} \wedge y_i=y_j\right\}\right|}{\left|\mathcal{N}(v_i)^{[n]}\right|}.
\end{equation}
It is obvious that $\mathit{HM}=\mathit{LC}\left(1\right)$ holds for a graph. 
From the Section \ref{sec:homo} and Table~\ref{tab:datasets_statistics} we can know that the $\mathit{LC}\left(1\right)$ of homophily graphs and heterophily graphs is significantly different. 
However, when we shift our focus from the first hop neighbors to a broader range, the $\mathit{LC}$ of both homophily graphs and heterophily graphs surprisingly exhibits the same characteristics. 
Specifically, for the 11 graph datasets (containing 8 homophily graphs and 3 heterophily graphs) used in this paper, we compute their $\mathit{LC}$ from the first hop neighbors to the fifth hop neighbor. 
The results are shown in Figure~\ref{fig:motivation}, where we can observe: 
\textbf{(1)} For both homophily graphs and heterophily graphs, as the distance between the neighbors and the anchor node increases, $\mathit{LC}$ shows a \textit{\textbf{diminishing trend}}. 
This means that closer nodes in both homophily graphs and heterophily graphs are more likely to have the same labels, {\it i.e.} they tend to be more similar, and vice versa.
\textbf{(2)} This diminishing trend is an \textbf{\textit{overall}} trend, {\it i.e.}, not all closer/further neighbors will be 100\% more/less similar.
Taking the Computer dataset as an example, about 20\% of the first hop neighbors have different labels from the anchor node, while about 60\% of the second hop neighbors have the same label as the anchor node, and these 20\% first hop neighbors should not be more similar to the anchor node than these 60\% second hop neighbors.

We call the above observations, which exist in both homophily graphs and heterophily graphs, the overall diminishing trend of label consistency. 
It provides us with a reliable means from the perspective of graph to judge the relative similarity and guide the learning process, {\it i.e.}, nodes that are closer in the graph are overall more similar than far-distant nodes.
We formalize the overall diminishing trend of label consistency as the following proposition:

\noindent\begin{proposition}
\label{proposition:ranking similarity}
For each node $v_i$ ({\it i.e.}, the anchor node) in a graph $\mathcal{G}$, the $n$-th hop neighbors of the anchor node $\mathcal{N}(v_i)^{[n]}$ are expected to be overall more similar to $v_i$ when compared to the nodes with a distance larger than $n$ hops from $v_i$.
Therefore, when the neighborhood range of $v_i$ we consider is $k$, for any $n \in [1,k]$, the $n$-th hop neighbors $\mathcal{N}(v_i)^{[n]}$ are expected to be overall more similar to the anchor node $v_i$ than the $(n+1)$-th hop neighbors $\mathcal{N}(v_i)^{[n+1]}$.
\end{proposition}

\begin{figure*}[!t]
\centering
\includegraphics[width=0.93\linewidth]{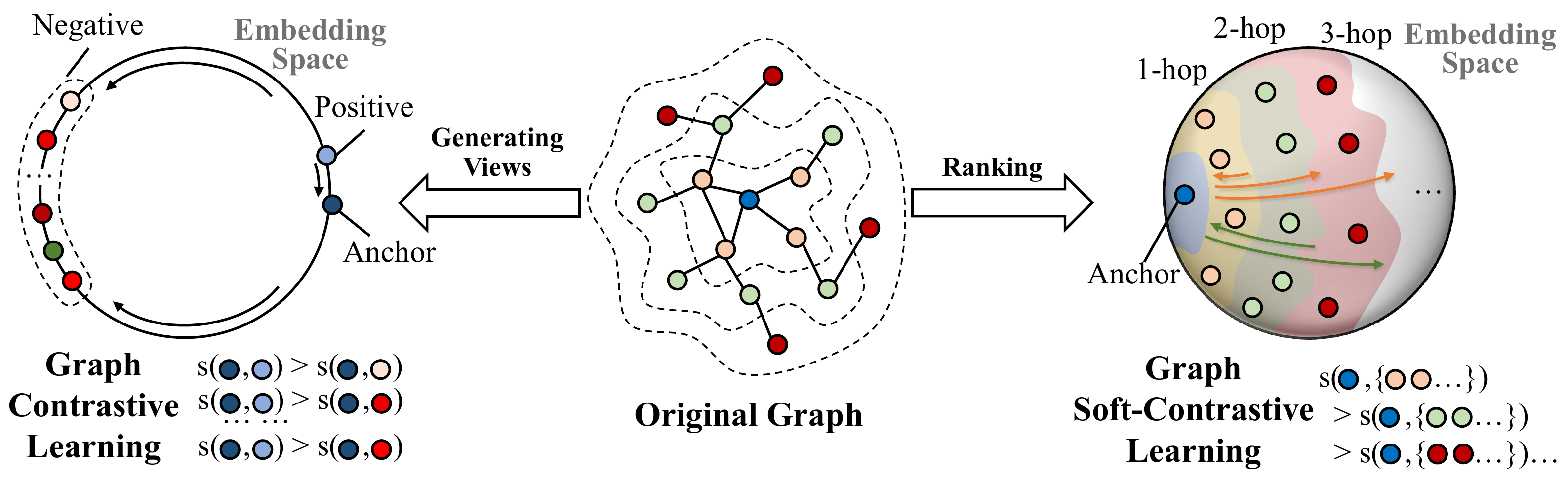}
\caption{A philosophy comparison between GCL and GSCL: (1) The light blue node and the dark blue node (left) are representations of the blue node (middle) in the two generated graph views, they are regarded as positive pairs and the distance between them will be reduced, and the distance between them and all other nodes will be increased. (2) Nodes of different colors (right) are the representations of different hop neighbors of blue nodes (middile), the soft margins represent an overall ranking trend that not all closer neighbors must be more similar than those further ones.}
\vspace{-4mm}
\label{fig:framework}
\end{figure*}

%% file: 4-method.tex
\section{Methodology}
In this section, we first introduce how to handling multiple positive examples robustly and give an overview of our proposed GSCL. 
Then, we present the paradigm of GSCL. 
Finally, we discuss the differences and connections of GSCL to current GCL methods.

\subsection{Handling Multiple Positive Examples Robustly}\label{sec:handling}

\subsubsection{\textbf{InfoNCE}}
We start with the most basic form and idea of GCL.
Classic GCL methods, like GRACE~\cite{Zhu2020DeepGC} and GCA~\cite{Zhu2021GraphCL}, firstly create two augmented views for a graph, then pull together the representation of the same node from the two graph views while pushing apart every other node pair to learn node representation (as shown in Figure \ref{fig:framework} left).
To achieve this goal, they optimize the classic InfoNCE loss~\cite{oord2018representation}:
\begin{equation}
\begin{aligned}
\mathcal{L}=-\log \frac{\exp \left(\frac{\theta(q, p)}{\tau}\right)}{\exp \left(\frac{\theta(q, p)}{\tau}\right)+\sum_{n \in \mathcal{N}} \exp \left(\frac{\theta(q, n)}{\tau}\right)},
\end{aligned}
\label{equation:infonce}
\end{equation}
where $\theta(\cdot,\cdot) = s(g(f(\cdot)), g(f(\cdot)))$ is a critic ($f(\cdot)$ is the encoder, $g(\cdot)$ is a nonlinear projection~\cite{Zhu2021GraphCL,chen2020simple,tschannen2019mutual} and $s(\cdot)$ is the cosine similarity), $\tau$ is a temperature parameter.
It is worth noting that for a query $q$, it has only \textbf{\textit{single}} positive example $p$ ({\it i.e.}, the same node from another view) yet a set of negative examples $\mathcal{N}$ ({\it i.e.}, all other nodes from both views).
However, in practice there must be more than one positive example, e.g., in graph, each node shares the same label with its many neighbors (as shown in Figure~\ref{fig:motivation}), but Equation~\ref{equation:infonce} will treat them all as negative examples.
So how can we handle \textbf{\textit{multiple}} positive examples $\mathcal{P}=\{p_{1},p_{2},\dots\}$ for per query $q$?

\subsubsection{\textbf{InfoNCE$\mathbf{_{out}}$}}
Supervised contrastive learning~\cite{khosla2020supervised} gives us a straightforward way to compute Equation~\ref{equation:infonce} for each of the positive examples from $\mathcal{P}$, {\it i.e.} take the sum over positive examples outside of the $\log$:
\begin{equation}
\begin{aligned}
\mathcal{L}^{\text {out }}=-\sum_{p \in \mathcal{P}} \log \frac{\exp \left(\frac{\theta(q, p)}{\tau}\right)}{\exp \left(\frac{\theta(q, p)}{\tau}\right)+\sum_{n \in \mathcal{N}} \exp \left(\frac{\theta(q, n)}{\tau}\right)},
\end{aligned}
\label{equation:out}
\end{equation}
Equation~\ref{equation:out} enforces similarity between $q$ and all positive examples during training, which requires the set of positive examples $\mathcal{P}$ to be clean~\cite{hoffmann2022ranking,khosla2020supervised}.
However, in the absence of supervised information, $\mathcal{P}$ can be noisy. 
For example, in homophily graph, although most of the first hop neighbors have the same label as the anchor node, it is not the case for all the first hop neighbors (as shown in Figure~\ref{fig:motivation} left).
So, it would be inappropriate to treat all the first hop neighbors as the set of positive examples for the anchor node when using InfoNCE$\mathbf{_{out}}$ to handle them.

\subsubsection{\textbf{InfoNCE$\mathbf{_{in}}$}}\label{sec:infonce_in}
MIL-NCE~\cite{miech2020end} gives us an alternative approach to handle multiple positive examples in an unsupervised setting, which is more robust to noisy positive examples, {\it i.e.} take the sum over positive examples inside the $\log$:
\begin{equation}
\begin{aligned}
\scalemath{0.95}{
\mathcal{L}^{\text {in }}=-\log \frac{\sum_{p \in \mathcal{P}} \exp \left(\frac{\theta(q, p)}{\tau}\right)}{\sum_{p \in \mathcal{P}} \exp \left(\frac{\theta(q, p)}{\tau}\right)+\sum_{n \in \mathcal{N}} \exp \left(\frac{\theta(q, n)}{\tau}\right)},}
\end{aligned}
\label{equation:in}
\end{equation}
Equation~\ref{equation:in} is not forced to set a higher similarity to all positive pairs and can neglect the noisy/false positives, given that a sufficiently large similarity is set for the true positives~\cite{hoffmann2022ranking,miech2020end}.
By optimizing this loss, the set of positive pairs ($q$ and $p \in \mathcal{P}$) will \textbf{\textit{overall}} be more similar than the set of negative pairs ($q$ and $n \in \mathcal{N}$) in the embedding space. 
So, Equation~\ref{equation:in} can give us a way to represent the overall \textbf{\textit{relative}} similarity, {\it i.e.}, how to make a set of pairs (the numerator in Equation~\ref{equation:in}) overall more similar than another set of pairs ($\sum_{n \in \mathcal{N}}$ in the denominator in Equation~\ref{equation:in}) in the embedding space.
We will use such property of InfoNCE$_{\mathrm{in}}$ in Section~\ref{sec:nr} to successfully model the overall diminishing trend of label consistency in the graphs.

\subsection{Framework Overview} 
GSCL aims to provide a new CL paradigm for graph data to avoid the limitation of current GCL by utilizing the structural properties of graphs. 
An overview of GSCL is presented in Figure~\ref{fig:framework}. 
GSCL takes a graph $\mathcal{G}$ as input, and the output is the learned representations. 
Specifically, GSCL includes three key steps: 
\textbf{(1)} Graph encoding: raw node features $\mathbf{X}$ are transformed into hidden representations $\mathbf{H}$ by the graph encoder $f$; 
\textbf{(2)} Neighborhood ranking preservation: for each node in the graph $\mathcal{G}$, we regard it as an anchor node, the ranking of similarities between the anchor node and the associated multi-hop neighbors within a neighborhood range is preserved following the Proposition \ref{proposition:ranking similarity}, with the developed pairwise and listwise ranking InfoNCE loss functions. 
\textbf{(3)} Optimization: the graph encoder $f$ is updated and optimized following the convention of gradient descent.

\subsection{Graph Soft-Contrastive Learning (GSCL)}
\label{section:GSCL}

\subsubsection{\textbf{Graph Encoder Backbone}} 
Following the convention of GCL, we take Graph Convolutional Network (GCN)~\cite{welling2016semi} as the graph encoder. 
Formally, the $l$-th layer of GCN can be represented as
\begin{equation}
    \text{GCN}_{l} (\mathbf{X}, \mathbf{A}) = \sigma(\hat{\mathbf{D}}^{-\frac{1}{2}} \hat{\mathbf{A}} \hat{\mathbf{D}}^{-\frac{1}{2}} \mathbf{X} \mathbf{W}_l), 
\end{equation}
where $\hat{\mathbf{A}}=\mathbf{A}+\mathbf{I}$ with $\mathbf{I}$ denotes the identity matrix, $\hat{\mathbf{D}}$ is the diagonal node degree matrix of $\hat{\mathbf{A}}$, $\sigma$ denotes the activation function, and $\mathbf{W}_l$ denotes the learnable weight matrix for the $l$-th layer. 
Suppose the graph encoder consists of $L$ layers of GCNs, then the graph encoder can be represented as 

\begin{equation}
\scalemath{0.95}{
\begin{aligned}
f(\mathbf{X}, \mathbf{A}) &= \text{GCN}_L(\text{GCN}_{L-1}(\cdots(\text{GCN}_{1} (\mathbf{X}, \mathbf{A}))\cdots, \mathbf{A}), \mathbf{A})\\ 
&=\mathbf{H}.
\end{aligned}
}
\label{equation:GCN}
\end{equation}

\subsubsection{\textbf{Neighborhood Ranking}}\label{sec:nr}
We expect the learned representations $\mathbf{H}$ to comply with Proposition \ref{proposition:ranking similarity}. 
For each node $v_i$ in the graph $\mathcal{G}$, the corresponding representation is denoted as $\mathbf{h}_i$ that is the $i$-th row of $\mathbf{H}$. 
When we consider the $k$-hop neighborhood range of node $v_i$, then the representations for the neighborhood can be denoted as:
\begin{equation}
    \mathbb{H}_{i}=\{\mathbb{H}^{[1]}_i, \mathbb{H}^{[2]}_i, \cdots, \mathbb{H}^{[k]}_i, \mathbb{H}^{[k+1]}_i\},
\end{equation} 
where $\mathbb{H}^{[n]}_i$ (for any $n \in [1,k]$) denotes the learned representations of the $n$-th hop neighbors $\mathcal{N}(v_i)^{[n]}$, $\mathbb{H}^{[k+1]}_i$ denotes the learned representations of all the nodes with a distance larger than $k$ hops from $v_i$.
Let $s(\cdot)$ denote the scoring function ({\it i.e.}, cosine similarity in our paper) to measure the similarity between a pair of node representations. 
Proposition \ref{proposition:ranking similarity} requires the learned node representations in the neighborhood to satisfy the following criterion:

\begin{equation}
\begin{aligned}
\sum_{\forall \mathbf{h}_\ast\in \mathbb{H}^{[1]}_i} &s(\mathbf{h}_i, \mathbf{h}_\ast)  >\sum_{\forall \mathbf{h}_\ast\in \mathbb{H}^{[2]}_i} s(\mathbf{h}_i, \mathbf{h}_\ast)>\cdots \\
&> \sum_{\forall \mathbf{h}_\ast\in \mathbb{H}^{[k]}_i} s(\mathbf{h}_i, \mathbf{h}_\ast)  >\sum_{\forall \mathbf{h}_\ast\in \mathbb{H}^{[k+1]}_i} s(\mathbf{h}_i, \mathbf{h}_\ast),
\end{aligned}
\label{equation:ranking}
\end{equation}
$\sum$ in Equation~\ref{equation:ranking} represents a generalized aggregation operation to represent the overall similarity of each hop neighbors to the anchor node, thus Equation~\ref{equation:ranking} reflects the overall diminishing trend of label consistency in graph data. 
To preserve such ranking relationships, we develop two new GCL loss functions in terms of pairwise and listwise ranking, respectively: 
\begin{itemize}
    \item \underline{Gated Pairwise Ranking InfoNCE (GSCL-Pairwise).}
    From the perspective of pairwise ranking~\cite{Burges2005LearningTR}, GSCL-Pairwise looks at a pair of neighbor sets from different hops at a time in the loss function.
    The neighborhood ranking can be illustrated as: given a pair of neighbor sets selected from two different hops, the neighbors from the smaller hop is expected to be overall more similar to the anchor node than the neighbors from the larger hop. 
    Then, we just need to iterate all possible pairs in the given $k$-hop neighborhood. 
    Inspired by RINCE~\cite{hoffmann2022ranking}, we develop the following gated pairwise ranking InfoNCE:
\end{itemize}
    \begin{flalign}
    \scalemath{0.7}{
        \mathcal{L} = - \sum\limits_{\mathbf{v}_i\in \mathcal{V}} \frac{1}{k} \sum\limits_{j=1}^{k} \sum\limits_{m=1}^{k-j+1} \log \left[ \min\{\frac{\sum\limits_{\mathbf{h}_\ast \in \mathbb{H}^{[j]}_i}\exp(\frac{\theta(\mathbf{h}_i, \mathbf{h}_\ast)}{\tau})}{\sum\limits_{\mathbf{h}_\ast \in \mathbb{H}^{[j]}_i}\exp(\frac{\theta(\mathbf{h}_i, \mathbf{h}_\ast)}{\tau}) + \sum\limits_{\mathbf{h}_\diamond \in \mathbb{H}^{[j+m]}_i}\exp(\frac{\theta(\mathbf{h}_i, \mathbf{h}_\diamond)}{\tau})}, \alpha \}\right], }
    \label{equation:pair-wise ranking infoNCE}
    \end{flalign}
\begin{itemize}
  \item[] where $\theta(\cdot,\cdot) = s(g(\cdot), g(\cdot))$ is a critic ($g(\cdot)$ is a nonlinear projection to enhance the expression power of the critic function~\cite{Zhu2021GraphCL,chen2020simple,tschannen2019mutual}, and it's implemented with a two-layer perceptron model here), $\tau$ denotes a temperature parameter, and $\alpha\in (0, 1)$ is the threshold to ensure a safe distance between pairs to avoid over-pushing two different-hop neighbors to extremely distant positions in the embedding space. It is worth noting that we use InfoNCE$\mathbf{_{in}}$ ({\it i.e.}, Equation~\ref{equation:in}) in a recursive manner to handle all neighbors from one smaller hop ({\it i.e.}, the positive examples in each iteration) robustly, thereby satisfying the overall nature of the diminishing trend described in Proposition \ref{proposition:ranking similarity}.
\end{itemize}

\begin{itemize}
    \item \underline{Gated Listwise Ranking InfoNCE (GSCL-Listwise).} 
    From the perspective of listwise ranking~\cite{cao2007learning}, GSCL-Listwise directly look at the neighbors from a given hop and the entire list of remaining neighbors from larger hops.
    The neighborhood ranking can be interpreted as: neighbors from a given hop is expected to be overall more similar to the anchor node than the remaining neighbors from larger hops. 
    We therefore need to take the given hop as the reference and enumerate all possible neighbors from larger hops. 
    Similarly, we develop the gated listwise ranking InfoNCE:
\end{itemize}

\begin{equation}
    \mathcal{L} = - \sum\limits_{\mathbf{v}_i\in \mathcal{V}} \frac{1}{k} \sum\limits_{j=1}^{k} \log \left[ \min \{ \frac{\sum\limits_{\mathbf{h}_\ast \in \mathbb{H}^{[j]}_i}\exp(\frac{\theta(\mathbf{h}_i, \mathbf{h}_\ast)}{\tau})}{\sum\limits_{j'=j}^{k+1} \sum\limits_{\mathbf{h}_\diamond \in \mathbb{H}^{[j']}_i} \exp(\frac{\theta(\mathbf{h}_i, \mathbf{h}_{\diamond})}{\tau}) },  \beta \}\right],
    \label{equation:list-wise ranking infoNCE}
\end{equation} 
\begin{itemize}
  \item[] where $\beta\in (0, 1)$ is the threshold to secure the listwise ranking in the reasonable range, which is similar to $\alpha$ in the gated pairwise ranking InfoNCE.
\end{itemize}

\subsubsection{\textbf{Optimization}} 
We optimize the graph encoder $f$ by minimizing the loss function $\mathcal{L}$. 
We exploit Gradient Stochastic Decent (GSD) to learn the optimal parameters of $f$.

\subsection{Neighborhood Sampling}
With more hops of neighbors considered, the number of nodes in the neighborhood would expand explosively, resulting in overwhelming computation. 
The situation is even exacerbated for large-scale graphs, which jeopardizes the applicability of the proposed GSCL. 
Therefore, we propose neighborhood sampling strategies to reduce the candidate neighborhood size. 
Specifically, we propose two of sampling strategies as follows:
\begin{itemize}
    \item \underline{Uniform Sampling.} We assign each node in the $n$-th hop neighbors $\mathcal{N}(v_i)^{[n]}$ with the same probability subject to the numbers of neighbors in the $n$-th hop as 
    \begin{equation}
        P(v_\ast) = \frac{1}{|\mathcal{N}(v_i)^{[n]}|}, \forall v_\ast \in \mathcal{N}(v_i)^{[n]}.
    \end{equation}
    \item \underline{Weighted Sampling Based on PageRank.} Instead of treating all neighbors from the same hop equally, we pay more attention to the nodes with higher importance. 
    Specifically, we first run PageRank~\cite{brin1998anatomy} over the graph $\mathcal{G}$ to assign each node a importance score. 
    Let $\text{PageRank}(v_\ast)$ denote the importance score of the node $v_\ast$. 
    We then assign the probability by normalizing the corresponding importance scores over the neighors from the same hop: 
    \begin{equation}
        P(v_\ast) = \frac{\text{PageRank}(v_\ast)}{\sum \limits_{v_\diamond \in \mathcal{N}(v_i)^{[n]}}\text{PageRank}(v_\diamond)} , \forall v_\ast \in \mathcal{N}(v_i)^{[n]}.
    \end{equation}
\end{itemize}

Based on the proposed sampling strategies, we sample a fixed size neighbors from each hop to construct a reduced neighborhood $\mathbb{\Tilde{H}}_{i}=\{\mathbb{\Tilde{H}}^{[1]}_i, \mathbb{\Tilde{H}}^{[2]}_i, \cdots, \mathbb{\Tilde{H}}^{[k]}_i,\mathbb{\Tilde{H}}^{[k+1]}_i\}$, where $\mathbb{\Tilde{H}}^{[n]}_i$ (for any $n\in[1,k]$) denotes the sampled candidate neighbors for the $n$-th hop. 
We then calculate the gated pairwise or listwise ranking InfoNCE based on the reduced neighborhood $\mathbb{\Tilde{H}}_{i}$ using Equation~\ref{equation:pair-wise ranking infoNCE} and Equation~\ref{equation:list-wise ranking infoNCE}, respectively.

\subsection{Comparison to Other GCL Methods}
\label{section:comparison to GCL} 
\subsubsection{\textbf{Different Settings}}
Table~\ref{table:setting comparison} presents the detailed comparison of settings between representative GCL methods and our proposed GSCL.  
Early GCL methods like GRACE~\cite{Zhu2020DeepGC} and GCA~\cite{Zhu2021GraphCL} use data augmentation to generate views for graphs, and regard the two augmented views of the same data example as positive pairs whereas views from different data examples as negative pairs.
In order to achieve good performance, these binary-contrastive methods often require a large number of negative samples, so some negative-sample-free approaches such as BGRL~\cite{thakoor2021large} and CCA-SSG~\cite{Zhang2021FromCC} try to use special loss functions and tricks to avoid negative samples.
But how to generate label-invariant and optimal views for a graph through data augmentation is non-trivial, data-augmentation-free methods like SUGRL~\cite{Mo2022SimpleUG} AFGRL~\cite{Lee2022AugmentationFreeSL} attempt to generate graph views without data augmentation.
Despite variations in motivations and implementations, the fundamental ideas underlying all these GCL methods are the same, they all aim to model invariance in an absolute manner, {\it i.e.}, specifying absolutely similar pairs in the graph views.
GSCL abandons such setting, but models similarity in a soft and relative manner by preserving the overall ranking relationship of similarities within neighborhoods.  
GSCL is conducted without views generation and any needs for graph augmentations, and its core idea is based on the structural information in the graph, {\it i.e.}, the overall diminishing trend of label consistency, thus avoiding the limits of traditional GCL.

\begin{table}[!t]
\scriptsize
\centering
\setlength\tabcolsep{2pt}
\caption{A setting comparison between state-of-the-art GCL methods and our proposed GSCL.}
\begin{tabular}{@{}lccc@{}}
\toprule
  & \textbf{ Data Augmentation } & \textbf{ Generating Views } & \textbf{ Absolute Similarity } \\
 \midrule
GRACE~\cite{Zhu2020DeepGC} &          {\color{green} \CheckmarkBold }              &         {\color{green} \CheckmarkBold }                     &        {\color{green} \CheckmarkBold }          \\
GCA~\cite{Zhu2021GraphCL} &           {\color{green} \CheckmarkBold }               &         {\color{green} \CheckmarkBold }                     &        {\color{green} \CheckmarkBold }          \\
BGRL~\cite{thakoor2021large} &           {\color{green} \CheckmarkBold }               &         {\color{green} \CheckmarkBold }                   &        {\color{green} \CheckmarkBold }           \\
CCA-SSG~\cite{Zhang2021FromCC} &           {\color{green} \CheckmarkBold }               &         {\color{green} \CheckmarkBold }                   &        {\color{green} \CheckmarkBold }           \\
SUGRL~\cite{Mo2022SimpleUG} &        {\color{red}   \XSolidBrush }              &         {\color{green} \CheckmarkBold }                   &    {\color{green} \CheckmarkBold }           \\
AFGRL~\cite{Lee2022AugmentationFreeSL}&           {\color{red}   \XSolidBrush }              &         {\color{green} \CheckmarkBold }                   &       {\color{green} \CheckmarkBold }        \\
OURS &            {\color{red}   \XSolidBrush }          &            {\color{red}   \XSolidBrush }             &    {\color{red}   \XSolidBrush }       \\ 
\bottomrule
\end{tabular}
\vspace{-0.30cm}
\label{table:setting comparison}
\end{table}

\begin{table*}[!t]
\centering
\small
\caption{The characteristics and statistics of datasets, where $\mathit{HM}$ represents the homophiliy metric, and ``\#$n$-hop neighbors" represents the average number of $n$-th hop neighbors.}
\label{tab:datasets_statistics}
\setlength{\tabcolsep}{1.80mm}{
\begin{tabular}{@{}lcrrrrrrrrrr@{}}
\toprule
\textbf{Dataset} & \textbf{Type} &  $\mathit{HM}$  & \textbf{\#Nodes} & \textbf{\#Edges} & \textbf{\#Features} & \textbf{\#Classes} & \textbf{\makecell[c]{\#1-hop \\ {\scriptsize neighbors}}} & \textbf{\makecell[c]{\#2-hop \\ {\scriptsize neighbors}}} & \textbf{\makecell[c]{\#3-hop \\ {\scriptsize neighbors}}} & \textbf{\makecell[c]{\#4-hop \\ {\scriptsize neighbors}}} & \textbf{\makecell[c]{\#5-hop \\ {\scriptsize neighbors}}} \\  \midrule
Cora                 & Homophiliy   &  0.8252  &       $2,708$           &       $10,556$           &        $1,433$             &         7           &  $3.90$   & $31.9$   &  $91.3$   & 244.9  &  438.4   
\\
Citeseer                 &  Homophiliy  &  0.7062  &     $3,327$            &       $9,228$           &         $3,703$            &        6            &  $2.7$   &  $11.4$  &  $28.4$   & 52.7   &  78.1    
\\
Pubmed                &  Homophiliy  & 0.7924   &      $19,717$          &       $88,651$           &         500            &            3        &  $4.5$   & $54.6$   &  334.6    & 1596.7  &  3321.4  
\\
WikiCS                 &  Homophiliy  &  0.6409  &    $11,701$           &        $216,123$          &        $300$             &        10            &  $24.8$  & $1051.4$ &  $4701.8$ & 3199.0 &  647.9  
\\
Computers                 &  Homophiliy  & 0.7853   &    $13,752$             &      $245,861$            &       $767$              &        10            &  $35.8$  & $1812.0$ &  $5644.3$ & 4332.6 &  1047.8  
\\
Photo                 & Homophiliy  &  0.8365  &    $7,650$           &        $119,081$          &          $745$           &          8          &  $31.1$  & $770.7$  &  $1716.5$ & 2162.3 &  1729.2 
\\
Co.CS                 & Homophiliy  &  0.8320  &    $18,333$           &       $81,894$           &          $6,805$           &       15             &  $8.9$   & $98.7$   &  $765.0$  & 3225.5 &  6158.7  
\\
ogbn-arxiv                 & Homophiliy  &  0.5532  &    $169,343$           &       $1,166,243$           &          $128$           &       40             &  $7.0$   & $2038.3$   &  $8234.6$  & 16160.1 &  19269.8  
\\
Chameleon                 &  Heterophily  &  0.2425  &    $2,277$           &       $36,101$           &          $2,325$           &       5             &  $27.6$   & $531.1$   &  $507.7$  & 761.6 &  298.7 
\\
Squirrel                 &  Heterophily  &  0.2175  &    $5,201$           &       $217,073$           &          $2,089$           &       5             &  $76.3$   & $1615.8$   &  $1846.5$  & 1171.4 &  413.1  
\\
Actor                 &  Heterophily  &  0.1848  &    $7,600$           &       $33,544$           &          $931$           &       5             &  $3.9$   & $96.7$   &  $482.1$  & 1444.4 &  1651.5 
\\
\bottomrule
\end{tabular}}
\vspace{-0.30cm}
\end{table*}

\subsubsection{\textbf{Complexity Analysis}} \label{section:Complexity_Analysis} 
Suppose a graph $\mathcal{G}$ includes $N$ nodes, and the average degree is $d$.
For fair comparison, as our proposed GSCL does not include graph augmentations, we only consider the times of calculating scoring function $s(\cdot)$ for evaluating the time complexity. 
For a typical GCL pipeline with the binary contrastive justification ({\it i.e.}, InfoNCE), we consider $Q_p$ positive samples and $Q_n$ samples for each node. 
Then, the time complexity of GCL is $\mathcal{O}(N(Q_p+Q_N))$. 
For our proposed GSCL, we consider the complexity in two cases: 
(1) in gated pairwise ranking InfoNCE loss, GSCL calculates $N\cdot k\sum\limits_{i=1}^{k+1}d^i$ times of $s(\cdot)$; 
(2) in gated listwise ranking InfoNCE loss, GSCL calculates $N(\sum\limits_{i=1}^{k}i\cdot d^i+k\cdot d^{k+1})$ times of $s(\cdot)$. 
In fact, following the raw format of Equation~\ref{equation:pair-wise ranking infoNCE} and Equation~\ref{equation:list-wise ranking infoNCE}, we repeatedly calculate the similarity between pairs. 
We can restore the calculated similarities in the memory to avoid duplicates, which means the actual time complexity of the pairwise and listwise cases is the same as is $\mathcal{O}(Nd(1+d+d^2))$. 
Moreover, suppose the sample size is $d'$ ($d \gg d'$) with considering the neighborhood sampling strategies, the complexity of GSCL can be further reduced to $\mathcal{O}(Nd'(1+d'+d'^2))$. 
Therefore, the complexity of GSCL is very closed to, even smaller than GCL by adjusting the sample size $d'$.

%% file: 5-experiment.tex
\section{Experiment}
In this section, extensive experiments are conducted to evaluate the effectiveness of the proposed  GSCL.
Our experiments aim to answer the following research questions (\textbf{RQ}): 
\begin{itemize}
\item \textbf{RQ1:} 
Does our proposed GSCL actually work?
Can GSCL outperform the traditional SOTA GCL methods on a variety of graphs of different sizes and properties? 
\item \textbf{RQ2:} How well does GSCL perform in other tasks besides node classification?
\item \textbf{RQ3:} How does the neighborhood range affect the performance of GSCL? Is this effect consistent across graph datasets with different sizes and types?
\item \textbf{RQ4:} Can GSCL with neighborhood sampling strategie work well when neighborhood size explosively expands?
\item \textbf{RQ5:} Is it really efficient and robust to use InfoNCE$_{\mathrm{in}}$ to design the loss functions of GSCL?
\item \textbf{RQ6:} How does the thresholds $\alpha$ and $\beta$ affect the effectiveness of GSCL?
\item \textbf{RQ7:} What is the performance of GSCL when scaling to a larger dimension?
\item \textbf{RQ8:} Can the proposed GSCL effectively fulfill the requirement of Proposition \ref{proposition:ranking similarity}? 
\end{itemize}

\subsection{Experimental Setting}
\subsubsection{\textbf{Datasets}} 
we conduct a comprehensive set of experiments on 11 real-world benchmark datasets that have been widely used in the graph community. 
The datasets consist of 8 homophiliy graphs, namely Cora, Citeseer, Pubmed, WikiCS, Amazon-Computers (Computers), Amazon-Photo (Photo), Coauthor-CS (Co.CS) and ogbn-arxiv, along with 3 heterophily graphs, namely Chameleon, Squirrel, and Actor. 
The characteristics and statistics of datasets are summarized in Table \ref{tab:datasets_statistics}, from which we can see that these datasets vary greatly in characteristic, size, feature, etc.
This is in order to fully validate the performance of our method in different scenarios.
Following are their detailed descriptions:
\begin{itemize}
\item \textbf{Cora}, \textbf{CiteSeer} and \textbf{PubMed} are 3 widely-used citation networks~\cite{sen2008collective}, where nodes correspond to papers and edges represent a citation link between two papers.
Each node contains a sparse bag-of-words feature and an article type label.
All three datasets have a public train/valid/test split, which we utilize directly.
\item \textbf{WikiCS} is built on the reference network in Wikipedia, where nodes represent articles about computer science and edges represent hyperlinks between these articles. 
Node feature is the average of the GloVe~\cite{pennington2014glove} word embeddings of all words in the article.
It provides 20 public train/valid/test splits, which we use directly.
\item \textbf{Amazon-Computers} and \textbf{Amazon-Photo} are 2 graphs constructed by Amazon~\cite{mcauley2015image}, where nodes represent products, and edges are constructed between them when they are often purchased together.
\textbf{Coauthor-CS} is derived from the MAG~\cite{sinha2015overview}, where nodes represent authors and edges indicate co-authorship.
As these datasets do not come with a standard split, we split them randomly into train, validation and test nodes (10\%/10\%/80\%).
\item \textbf{ogbn-arxiv} is a directed graph which represents the citation network between all Computer Science (CS) arXiv papers~\cite{hu2020open}.
Each node represents an arXiv paper and each directed edge indicates that one paper cites another one.
We directly use the official split.
\item \textbf{Chameleon} and \textbf{Squirrel} are Wikipedia page-page networks on 2 different topics, where nodes represent articles and edges reflect mutual links between them~\cite{rozemberczki2021multi}.
\textbf{Actor} is the actor-only induced subgraph of the film-director-actor-writer network, where nodes correspond to actors, and edges denote co-occurrence of two actors on the same Wikipedia page~\cite{Pei2020GeomGCNGG}.
For these 3 datasets, we use the datasets and public splits provided by Geom-GCN~\cite{Pei2020GeomGCNGG}.
\end{itemize}

\begin{table*}[!t]
    \centering
    \caption{Classification accuracies on 7 GCL benchmark datasets (mean$\pm$std), the larger value implies a better classification result. `Ave. Rank' refers to the average ranking of a method across the 7 datasets, the smaller value of which indicates better overall performance. \textbf{Bold} indicates the best performance and \underline{underline} indicates the runner-up. $\mathbf{X}$ represents the node features, $\mathbf{A}$ represents the adjacency matrix, $\mathbf{Y}$ represents the node labels, and `OOM' represents out of memory on a 32GB GPU.}
    \label{tab:main_results}
    \setlength\tabcolsep{3pt}
    \begin{threeparttable}
    \setlength{\tabcolsep}{1.2mm}{
    \begin{tabular}{@{}lcccccccc|c@{}}
        \toprule
        \textbf{Method} &   \textbf{Training Data}                  &   \textbf{Cora}   &   \textbf{Citeseer}   &   \textbf{Pubmed} &   \textbf{WikiCS}     &   \textbf{Computers}  &   \textbf{Photo}      &   \textbf{Co.CS} &  \textbf{Ave. Rank}    \\ 
        \midrule
        GCN             &   $\mathbf{X}, \mathbf{A}, \mathbf{Y}$    &   $81.5$          &   $70.3$              &   $79.0$          &   $77.19 \pm 0.12$    &   $86.51 \pm 0.54$    &   $92.42 \pm 0.22$    &   $93.03 \pm 0.31$ &  14.4  \\
        GAT             &   $\mathbf{X}, \mathbf{A}, \mathbf{Y}$    &   $83.0 \pm 0.7$  &   $72.5 \pm 0.7$      &   $79.0 \pm 0.3$  &   $77.65 \pm 0.11$    &   $86.93 \pm 0.29$    &   $92.56 \pm 0.35$    &   $92.31 \pm 0.24$ &   11.6 \\
        \midrule
        Raw features       &   $\mathbf{X}$                            &   $47.9 \pm 0.4$  &   $49.3 \pm 0.2$      &   $69.1 \pm 0.3$  &   $71.98 \pm 0.00$    &   $73.81 \pm 0.00$    &   $78.53 \pm 0.00$    &   $90.37 \pm 0.00$ &    21.9\\
        Node2vec        &   $\mathbf{A}$                            &   $71.1 \pm 0.9$         &   $47.3 \pm 0.8$            &   $66.2 \pm 0.9$         &    $71.79 \pm 0.05$    &   $84.39 \pm 0.08$    &   $89.67 \pm 0.12$    &   $85.08 \pm 0.03$ &  21.4  \\

        DeepWalk        &   $\mathbf{A}$                            &   $70.7 \pm 0.6$          &   $51.4 \pm 0.5$              &   $74.3 \pm 0.9$          &   $74.35 \pm 0.06$    &   $85.68 \pm 0.06$    &   $89.44 \pm 0.11$    &   $84.61 \pm 0.22$ &  20.6  \\
        \midrule
        GAE             &   $\mathbf{X}, \mathbf{A}$                &   $71.5 \pm 0.4$  &   $65.8 \pm 0.4$      &   $72.1 \pm 0.5$  &   $70.15 \pm 0.01$    &   $85.27 \pm 0.19$    &   $91.62 \pm 0.13$    &   $90.01 \pm 0.71$ &  19.9  \\
        VGAE            &   $\mathbf{X}, \mathbf{A}$                &   $76.3 \pm 0.2$  &   $66.8 \pm 0.2$      &   $75.8 \pm 0.4$  &   $75.63 \pm 0.19$    &   $86.37 \pm 0.21$    &   $92.20 \pm 0.11$    &   $92.11 \pm 0.09$ &   17.1 \\ 
        GraphMAE        &   $\mathbf{X}, \mathbf{A}$                &   $84.2 \pm 0.4$  &   $73.4 \pm 0.4$      &   $81.1 \pm 0.4$  &   $77.12 \pm 0.30$                   &   $79.44 \pm 0.48$                   &   $90.71 \pm 0.40$                  &   $93.13 \pm 0.15$       &     10.3       \\
        DGI             &   $\mathbf{X}, \mathbf{A}$                &   $82.3 \pm 0.6$  &   $71.8 \pm 0.7$      &   $76.8 \pm 0.6$  &   $75.35 \pm 0.14$    &   $83.95 \pm 0.47$    &   $91.61 \pm 0.22$    &   $92.15 \pm 0.63$ &  16.7  \\
        MVGRL          &   $\mathbf{X}, \mathbf{A}$    &   $83.5 \pm 0.4$  &   $73.3 \pm 0.5$      &   $80.1 \pm 0.7$  &   $77.52 \pm 0.08$    &   $87.52 \pm 0.11$    &   $91.74 \pm 0.07$    &   $92.11 \pm 0.12$   & 11.0 \\
        GRACE         &   $\mathbf{X}, \mathbf{A}$                &   $81.9 \pm 0.4$  &   $71.2 \pm 0.5$      &   $80.6 \pm 0.4$  &   $78.19 \pm 0.01$    &   $86.25 \pm 0.25$    &   $92.15 \pm 0.24$    &   $92.93 \pm 0.01$   & 13.1  \\
        GCA             &   $\mathbf{X}, \mathbf{A}$                &   $82.1 \pm 0.1$  &   $71.3 \pm  0.4$     &   $80.2 \pm 0.4$  &   $78.30 \pm 0.00$    &   $87.85 \pm 0.31$    &   $92.49 \pm 0.09$    &   $93.10 \pm 0.01$    & 11.3  \\
        BGRL           &   $\mathbf{X}, \mathbf{A}$                &   $82.7 \pm 0.6$  &   $71.1 \pm 0.8$      &   $79.6 \pm 0.5$  &   $79.31 \pm 0.55$    &   $89.62 \pm 0.37$    &   $93.07 \pm 0.34$    &   $92.67 \pm 0.21$    & 10.0  \\
        G-BT            &   $\mathbf{X}, \mathbf{A}$                &   $81.5 \pm 0.4$               &   $71.9 \pm 0.5$                 &   $80.4 \pm 0.6$               &   $76.65 \pm 0.62$    &   $88.14 \pm 0.33$    &   $92.63 \pm 0.44$    &   $92.95 \pm 0.17$     &  11.9 \\   
        CCA-SSG         &   $\mathbf{X}, \mathbf{A}$                &   $84.2 \pm 0.4$  &   $73.1 \pm 0.3$      &   $81.6 \pm 0.4$  &   $78.65 \pm 0.14$                   &   $88.74 \pm 0.28$    &   $93.14 \pm 0.14$    &   $93.31 \pm 0.22$    &  5.0  \\
        GMI             &   $\mathbf{X}, \mathbf{A}$                &   $83.0 \pm 0.3$  &   $72.4 \pm 0.1$      &   $79.9 \pm 0.2$  &   $74.85 \pm 0.08$    &   $82.21 \pm 0.31$    &   $90.68 \pm 0.17$    &   OOM        &     15.7    \\
        COSTA           &   $\mathbf{X}, \mathbf{A}$                &   $83.3 \pm 0.3$               &   $72.1 \pm 0.3$                   &   $81.1 \pm 0.2$               &   $79.12 \pm 0.02$    &   $88.32 \pm 0.03$    &   $92.56 \pm 0.45$    &   $92.94 \pm 0.10$   & 7.9  \\
        SUGRL           &   $\mathbf{X}, \mathbf{A}$                &   $83.4 \pm 0.5$  &   $73.0 \pm 0.4$      &   $81.9 \pm 0.3$  &   $78.97 \pm 0.22$                  &   $88.91 \pm 0.22$      &   $92.85 \pm 0.24$      &   $92.83 \pm 0.23$           &   6.4     \\ 
        AFGRL           &   $\mathbf{X}, \mathbf{A}$                &   $83.2 \pm 0.4$               &   $72.6 \pm 0.3$                   &   $80.8 \pm 0.6$               &   $77.62 \pm 0.49$    &   $89.88 \pm 0.33$    &   $93.22 \pm 0.28$    &   $93.27 \pm 0.17$  
  & 6.6 \\         
        AF-GCL           &   $\mathbf{X}, \mathbf{A}$                &   $83.2 \pm 0.2$  &   $72.0 \pm 0.4$     &   $79.1 \pm 0.8$  &   $79.01 \pm 0.51$                &   $89.68 \pm 0.19$     &   $92.49 \pm 0.31$     &   $91.92 \pm 0.10$         &   10.4     \\ 
        AUTOSSL             &   $\mathbf{X}, \mathbf{A}$                &   $83.1 \pm 0.4$  &   $72.1 \pm 0.4$     &   $80.9 \pm 0.6$  &   $76.80 \pm 0.13$    &   $88.18 \pm 0.43$    &   $92.71 \pm 0.32$    &   $93.35 \pm 0.09$    &  8.9 \\   
        \midrule
       \rowcolor{gray!20} GSCL-Pairwise   &   $\mathbf{X}, \mathbf{A}$      &  \underline{$84.4 \pm 0.2$}  &   $\mathbf{73.7 \pm 0.4}$     &   \underline{$82.2 \pm 0.3$}  &   \underline{$80.14 \pm 0.51$}    &   $\mathbf{90.14 \pm  0.35}$    &   $\mathbf{93.42 \pm 0.44}$    &   $\mathbf{93.53 \pm 0.12}$  &  $\mathbf{1.4}$ \\
       \rowcolor{gray!20}  GSCL-Listwise   &   $\mathbf{X}, \mathbf{A}$                &   $\mathbf{84.5 \pm 0.4}$  &   \underline{$73.6 \pm 0.5$}      &   $\mathbf{82.7 \pm 0.4}$  &   $\mathbf{80.16 \pm 0.58}$    &   \underline{$89.99 \pm  0.38$}    &   \underline{$93.40 \pm 0.48$}    &   \underline{$93.50 \pm 0.12$}    &  \underline{1.6} \\
        \bottomrule
    \end{tabular}}
    \end{threeparttable}
    \vspace{-0.25cm}
\end{table*}

\begin{table}[]
\centering
\caption{Performance on 3 heterophily graphs measured in accuracy (mean$\pm$std).}
\label{tab:heter}
\begin{tabular}{@{}lccc@{}}
\toprule
\textbf{Method} & \textbf{Chameleon} & \textbf{Squirrel} & \textbf{Actor} \\ \midrule
    GAE            &        $39.13 \pm 1.34$            &        $34.65 \pm 0.81$           &       $25.36 \pm 0.23$         \\
    VGAE            &       $42.65 \pm 1.27$             &      $35.11 \pm 0.92$             &      $28.43 \pm 0.57$          \\
    DGI            &       $60.27 \pm 0.70$             &       $42.22 \pm 0.63$            &      $28.30 \pm 0.76$          \\
    GMI            &       $52.81 \pm 0.63$             &       $35.25 \pm 1.21$            &       $27.28 \pm 0.87$         \\
    GRACE            &       $61.24 \pm 0.53$             &      $41.09 \pm 0.85$             &       $28.27 \pm 0.43$         \\
    GCA            &        $60.94 \pm 0.81$            &      $41.53 \pm 1.09$             &      $28.89 \pm 0.50$          \\
    BGRL            &        $64.86 \pm 0.63$            &      $46.24 \pm 0.70$             &      $28.80 \pm 0.54$          \\
    AFGRL            &        $59.03 \pm 0.78$            &      $42.36 \pm 0.40$             &      $27.43 \pm 1.31$          \\
    AF-GCL            &      $65.28 \pm 0.53$              &      $52.10 \pm 0.67$             &     $28.94 \pm 0.69$           \\ \midrule
    \rowcolor{gray!20} GSCL-Pairwise            &        $\mathbf{69.25 \pm 0.89}$            &      $\mathbf{57.67 \pm 0.96}$             &         $\mathbf{30.32 \pm 0.91}$       \\
    \rowcolor{gray!20} GSCL-Listwise            &         \underline{$69.06 \pm 0.86$}           &       \underline{$57.53 \pm 0.92$}            &        \underline{$30.23 \pm 0.88$}        \\
 \bottomrule
\end{tabular}
\end{table}

From Table \ref{tab:datasets_statistics}, we can observe that starting from the 5-th hop neighbors, the average number of neighbors begins to decrease or increase by a smaller amount.
Furthermore, from Figure~\ref{fig:motivation} we can see that when the hop is greater than 4, the average proportion of neighbours with the same label as the anchor node becomes particularly small.
Therefore, the amount of similar neighbors at distances greater than 5 hops is especially small, so the neighborhood range $k$ ({\it i.e.}, the $k$ in Equation~\ref{equation:pair-wise ranking infoNCE} and \ref{equation:list-wise ranking infoNCE}) we choose for ranking in our experiments is from 1 to 4.

\subsubsection{\textbf{Baselines}}
To fully validate the performance of GSCL, we compare GSCL with 20 baselines, 
including 2 supervised methods ({\it i.e.}, GCN~\cite{welling2016semi} and GAT~\cite{Velickovic2018GraphAN}), 
2 random walk based methods ({\it i.e.}, Node2vec~\cite{grover2016node2vec} and DeepWalk~\cite{perozzi2014deepwalk}),
3 autoencoder based methods ({\it i.e.}, GAE, VGAE~\cite{Kipf2016VariationalGA} and GraphMAE~\cite{Hou2022GraphMAESM}),
7 augmentation-based GCL methods ({\it i.e.}, DGI~\cite{Velickovic2019DeepGI}, MVGRL~\cite{hassani2020contrastive}, GRACE~\cite{Zhu2020DeepGC}, GCA~\cite{Zhu2021GraphCL}, BGRL~\cite{thakoor2021large}, G-BT~\cite{bielak2022graph}), CCA-SSG~\cite{Zhang2021FromCC}, 
5 augmentation-free GCL methods ({\it i.e.}, GMI~\cite{Peng2020GraphRL}, COSTA~\cite{Zhang2022COSTACF}, SUGRL~\cite{Mo2022SimpleUG}, AFGRL~\cite{Lee2022AugmentationFreeSL} and AF-GCL~\cite{wang2022augmentation}),
and 1 multi-task self-supervised learning based method AUTOSSL~\cite{jin2021automated}.

\subsubsection{\textbf{Evaluation Protocol}} 
We follow the linear evaluation scheme adopted by previous works~\cite{Velickovic2019DeepGI, Zhu2020DeepGC, thakoor2021large, Zhang2021FromCC}.
Specifically, we first train the graph encoder (we employ GCN~\cite{welling2016semi} as the encoder for all methods) on all nodes in the graph without supervision by using the method described in the Section~\ref{section:GSCL}.
Following this, we freeze the GCN model and apply it to generate all nodes' embeddings.
Lastly, we use nodes' embeddings from the training set to train a logistic regression classifier for node classification, and report the classification accuracy on testing set when the performance on validation set gives the best result.

\subsubsection{\textbf{Implementation Details}}
GSCL is implemented in Pytorch and all experiments are performed on a NVIDIA V100 GPU with 32 GB memory.
All parameters of the GSCL and logistic regression classifier are optimized by the Adam optimizer~\cite{kingma2014adam}.
For fair evaluation, following previous works~\cite{Zhang2021FromCC, Lee2022AugmentationFreeSL, wang2022augmentation}, we report the average test accuracy with the corresponding standard deviation through 20 random initializations on each dataset. 
A few baselines do not use the public split of Cora, CiteSeer and PubMed, or do not provide results for some of the datasets,  we reproduce them based on their codes in the same experimental setup as ours.

\begin{table}[]
\centering
\caption{Performance on ogbn-arxiv dataset measured in accuracy (mean$\pm$std). }
\label{tab:ogb}
\begin{tabular}{@{}lccc@{}}
\toprule
\textbf{Method} & \textbf{Validation} & \textbf{Test}  \\ \midrule
    MLP            &        $57.65 \pm 0.12$            &        $55.50 \pm 0.23$                   \\
    node2vec            &       $71.29 \pm 0.13$             &      $70.07 \pm 0.13$                     \\ 
    Random-Init            &       $69.90 \pm 0.11$             &      $68.94 \pm 0.15$                     \\
    DGI            &       $71.26 \pm 0.11$             &       $70.34 \pm 0.16$                     \\
    GRACE-Subsampling            &       $72.61 \pm 0.15$             &       $71.51 \pm 0.11$                  \\
    G-BT            &       $71.16 \pm 0.14$             &      $70.12 \pm 0.18$                    \\
    BGRL            &      $72.53 \pm 0.09$              &       $71.64 \pm 0.12$                     \\ 
    CCA-SSG            &      $72.34 \pm 0.21$              &       $71.24 \pm 0.20$                     \\ 
    \midrule
    \rowcolor{gray!20} GSCL-Pairwise            &        \underline{$72.69 \pm 0.14$}                       &         \underline{$72.06 \pm 0.20$}       \\ 
    \rowcolor{gray!20} GSCL-Listwise            &         $\mathbf{72.75 \pm 0.12}$                     &        $\mathbf{72.24 \pm 0.19}$        \\
     \midrule
     Supervised GCN            &        $73.00 \pm 0.17$            &        $71.74 \pm 0.29$                   \\ 
    \bottomrule
\end{tabular}
\end{table}

We use grid search to find the optimal hyperparameters for each dataset. 
For generic hyperparameters, we find the optimal embedding dimension among $\{128, 256, 512, 1024\}$, the number of GCN layers among $\{1, 2\}$, the activation function of GCN among \{relu, prelu, rrelu\}, the learning rate and weight decay between $[1e^{-8}, 1e^{-2}]$, the temperature $\tau$ among $ \{0.1, 0.2, ..., 0.9\}$, the spacing size of $\tau$ between two adjoining hops among $ \{0, 0.0125, 0.025, 0.05, 0.1 \}$~\cite{hoffmann2022ranking}. 
For hyperparameters specific to the list-wise and pair-wise ranking loss, we find the optimal neighborhood range $k$ for ranking among $ \{1, 2, 3, 4\}$, the thresholds $\alpha$ and $\beta$ in $[0.0001, 1.0]$.

\subsection{Overall Performance on a Variety of Graphs (RQ1)}
\subsubsection{\textbf{Results on Homophiliy GCL Benchmarks}}
We first report the results of node classification task on 7 homophiliy graphs (the widely-used GCL benchmark) in Table \ref{tab:main_results}.
From the table, we have the following observations:
\textbf{(1)} Without relying on any data augmentations or views generation, both GSCL-Pairwise and GSCL-Listwise and can achieve superior performance across 7 datasets compared with 20 baselines ({\it i.e.}, outperforming all baselines and obtaining average rank of 1.6 and 1.4 respectively compared to the best average rank of 5.0 in the baselines), demonstrating the effectiveness and generalization of our simple ranking based soft-CL paradigm on a wide variety of homophiliy graphs.
\textbf{(2)} The performance of GSCL-Pairwise and GSCL-Listwise is close, they differ only by 0.2 in average ranks and achieve 3 and 4 SOTAs in the 7 datasets respectively. 
The similar good performance of these two different ranking implementations is a sufficient evidence that solving GCL via ranking within neighborhoods 

is feasible.
\textbf{(3)} Both SUGRL and AFGRL exploit the idea of using the first hop neighbors as positive examples~\cite{Mo2022SimpleUG, Lee2022AugmentationFreeSL}, which is actually a special case of our proposed GSCL when neighborhood range $k$ for ranking is 1.
GSCL's better results over them illustrates the importance of performing ranking in a broader neighborhood range for homophiliy graphs. 
\textbf{(4)} AUTOSSL directly make use of homophiliy as the guidance to effectively search multiple self-supervised pretext tasks~\cite{jin2021automated}. 
Our GSCL outperforms AUTOSSL on all datasets, suggesting that by utilizing more appropriate graph structural information, even just a single self-supervised pretext task can achieve sufficiently good downstream task performance.

\subsubsection{\textbf{Results on Heterophily Graphs}}
We further evaluate the performance of GSCL on 3 heterophily graphs that employed in ~\cite{Pei2020GeomGCNGG,wang2022augmentation}, and the results are presented in Table \ref{tab:heter} (the results of baselines are taken from~\cite{wang2022augmentation}).
From the table, we make several observations:
\textbf{(1)} Both GSCL-Pairwise and GSCL-Listwise substantially outperform all baselines on all 3 datasets, thus validating that our method, {\it i.e.}, modeling the overall diminishing trend of label consistency, is also appropriate and effective on heterophily graphs.
\textbf{(2)} GSCL-Pairwise slightly outperforms GSCL-Listwise on all 3 datasets, probably because most of the nodes in heterophily graphs are dissimilar (as shown in Figure \ref{fig:motivation} right), and GSCL-Pairwise ranks the similarity of neighbors in a more fine-grained manner ({\it i.e.}, in pairs) than GSCL-Listwise, thus GSCL-Pairwise is better suited to heterophily graphs.
\textbf{(3)} AFGRL~\cite{Lee2022AugmentationFreeSL}  treats the first hop neighbors as positive examples (similar nodes) thus implicitly modeling homophiliy. 
The worse performance of AFGRL compared to other GCL methods suggests that modeling homophiliy on heterophily graphs is harmful.

\subsubsection{\textbf{Results on Large-Scale OGB Dataset}}
Finally, we investigate the effectiveness and scalability of GSCL on a large-scale graph dataset ogbn-arxiv from OGB benchmark~\cite{hu2020open}. 
We expand the graph encoder of GSCL to use 3 GCN layers and adopt sub-sampling for negative examples following previous works~\cite{hu2020open, thakoor2021large}. 
We report both the validation and test accuracy of baselines and GSCL in Table \ref{tab:ogb}, as is a convention for this task since the dataset is split based on a chronological ordering.
As shown in the table, both GSCL-Pairwise and GSCL-Listwise outperform all other unsupervised baselines, thus validating the effectiveness and scalability of GSCL on the large-scale graph.

\subsubsection{\textbf{Brief Summary}}
From the above experimental results, it can be concluded that by modeling the overall diminishing trend of label consistency in the graphs, {\it i.e.}, ranking neighbors within a neighborhood range, GSCL can consistently achieve superior performance on homophiliy, heterophily and large-scale graphs compared to various baselines.
Thereby empirically verifying the rationality, effectiveness, generalization and scalability of GSCL.

\begin{table}[!t]
\centering
\caption{Performance of node clustering in terms of NMI and performance of node similarity search in terms of Sim@5 ({\it i.e.}, average ratio among 5 nearest neighbors sharing the same label as the query node).}
\label{tab:other}
\setlength{\tabcolsep}{1.8mm}{
\begin{tabular}{@{}lcccccc@{}}
\toprule
\textbf{Dataset} & \multicolumn{2}{c}{\textbf{WikiCS}} & \multicolumn{2}{c}{\textbf{Computers}} & \multicolumn{2}{c}{\textbf{Photo}} \\ 
\textbf{Metric}  & \textbf{NMI}    & \textbf{Sim@5}    & \textbf{NMI}      & \textbf{Sim@5}     & \textbf{NMI}    & \textbf{Sim@5}   \\ \midrule
GRACE                 &       0.4282          &    0.7754               &      0.4793             &          0.8738          &      0.6513           &      0.9155            \\
GCA                 &        0.3373         &    0.7786               &          0.5278         &           0.8826        &       0.6443          &     0.9112             \\
BGRL                 &       0.3969          &     0.7739              &       0.5364            &         0.8947           &       0.6841          &     0.9245             \\
AFGRL                 &       0.4132          &           0.7811        &         0.5520          &     0.8966               &      0.6563           &      0.9236            \\ \midrule
\rowcolor{gray!20} GSCL-Pairwise                 & \underline{0.4354}                 &      \underline{0.7919}             &         \textbf{0.5643}          &         \textbf{0.8992}           &      \textbf{0.6935}          &        \textbf{0.9302}         \\
\rowcolor{gray!20} GSCL-Listwise                 & \textbf{0.4376}                 &   \textbf{0.7951}                &        \underline{0.5627}           &       \underline{0.8988}             &       \underline{0.6917}          &         \underline{0.9297}         \\ \bottomrule
\end{tabular}}
\end{table}

\subsection{Performance on Other Tasks (RQ2)}
To verify the ability of GSCL on other tasks, we also evaluate the performance of node clustering and node similarity search on WikiCS, Computers, and Photo datasets.
For these two tasks, we perform evaluations on the learned embeddings at every epoch and report the best performance following previous work~\cite{Lee2022AugmentationFreeSL},  and the results are presented in Table \ref{tab:other} (the results of baselines are taken from~\cite{Lee2022AugmentationFreeSL}).
It can be observed from the table that on all three datasets, both GSCL-Pairwise and GSCL-Listwise generally outperform other baselines in both node clustering and node similarity search tasks.
This indicates that by modeling the overall diminishing trend of label consistency in the graphs, GSCL can not only solve common node classification task, but also achieve excellent performance on other tasks, thus demonstrating the excellent task generalization ability of GSCL for graph-structured data.
It is also worth noting that AFGRL exploits the idea of using the first hop neighbors as positive examples~\cite{Lee2022AugmentationFreeSL}, which is actually a special case of our proposed GSCL when neighborhood range $k$ for ranking is 1.
GSCL generally outperforms AFGRL by a substantial margin, indicating that expanding the range of graph-structured information (from first hop neighbors to a broader neighborhood range) is beneficial for solving problems of graph-structured data, especially for clustering task in which the global structural information is crucial~\cite{Lee2022AugmentationFreeSL}.

\begin{figure*}[!h]
\centering
\includegraphics[width=1.0\linewidth]{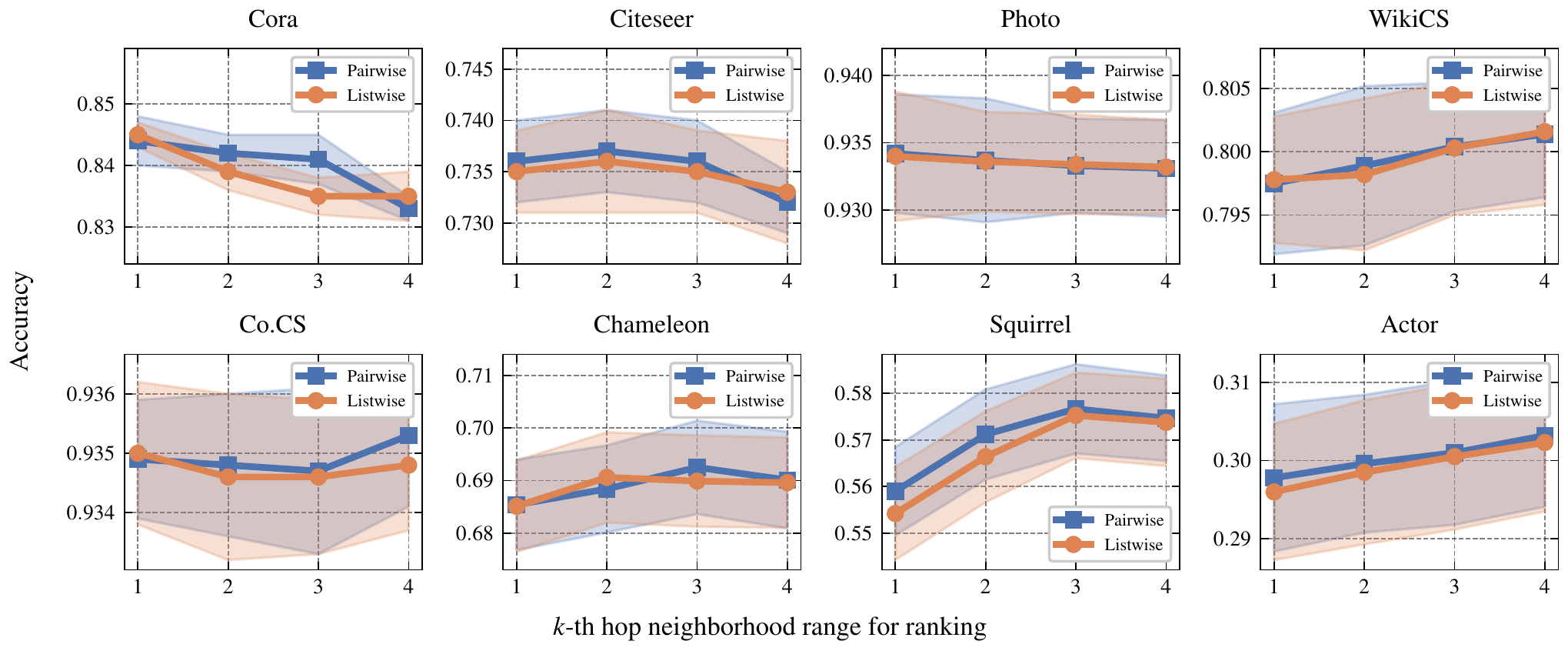}
\caption{Results of GSCL with neighborhood range changing as $k=1, 2, 3, 4$, respectively.}
\label{fig:different_hops}
\end{figure*}

\begin{figure}[]
\centering
\includegraphics[width=1.0\linewidth]{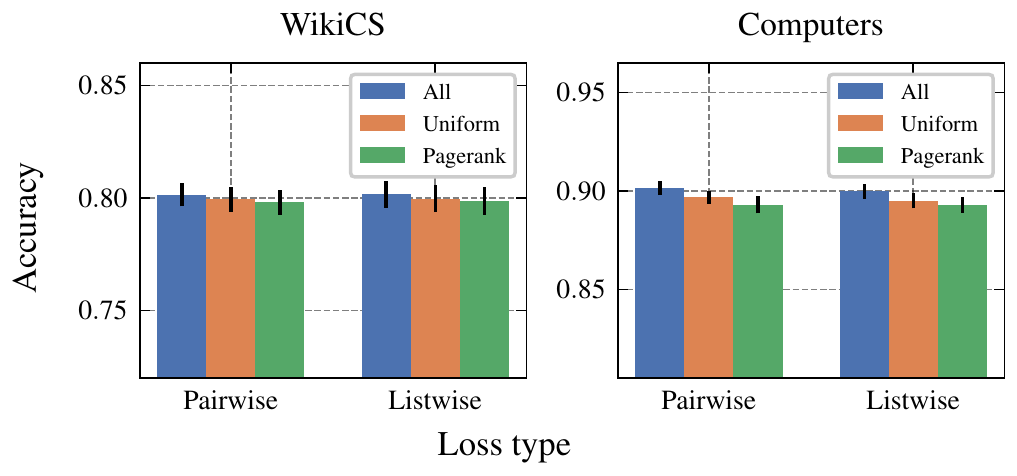}
\caption{Results of GSCL incorporated with different neighborhood sampling strategies.}
\label{fig:different_sampling_strategies} 
\end{figure}

\subsection{Study of $k$-hop Neighborhood Range for Ranking (RQ3)} 
For each node in the graph, we are ranking the neighbors within a neighborhood range of $k$ in which they are located, and we now investigate the effects of $k$ on GSCL's performance.
We conduct experiments on Cora, CiteSeer, Photo (\textit{small homophiliy graphs}), WikiCS, Co.CS (\textit{large homophiliy graphs}), Chameleon, Squirrel and Actor (\textit{heterophily graphs}) to investigate the effect of neighborhood range on GSCL-Listwise and GSCL-Pairwise respectively. 
As shown in Figure~\ref{fig:different_hops}, we can observe that the best performance of GSCL is achieved at different $k$-hop neighborhood range due to different characteristics of datasets:
\textbf{(1)} On small homophiliy graphs, GSCL achieves the most accurate results with $k=1$ or $2$, but the performance will decrease with $k$ going up.
\textbf{(2)} On large homophiliy graphs and heterophily graphs, the results of GSCL overall improve as $k$ increases and the highest results for GSCL are always obtained when $k=3$ or $4$.
Such results indicate that for small homophiliy graphs, a local neighborhood range is sufficient to obtain excellent performance, while for large homophiliy graph and heterophily graphs, a larger neighborhood range is often required.
The reasons may be:
\textbf{(1)} For homophiliy graphs, as shown in Figure~\ref{fig:motivation} left that the label consistency decreases substantially as $n$ increases, reaching around 20\% when $n=4$.
So when the size of a homophiliy graph is small, the number of third or forth hop neighbors with the same label as the anchor nodes will be extremely small, ranking within such range is unnecessary and may even lead to negative effects.
\textbf{(2)} For heterophily graphs, as shown in Figure~\ref{fig:motivation} right that their label consistency is small overall and the difference of label consistency between adjacent hops is not large.
Thus for heterophily graphs, it is important to expand the neighborhood range for ranking in order to perform a more sufficient ranking and comparison, which will lead to a better performance.
Moreover for most graph datasets, GSCL can achieve the best performance before the forth hop neighborhood.
This is because GSCL treats all nodes in the graph $\mathcal{G}$ as anchor nodes ($\sum{\mathbf{v}_i\in\mathcal{V}}$ in Equation \ref{equation:pair-wise ranking infoNCE} and Equation \ref{equation:list-wise ranking infoNCE}), setting the neighborhood range within 4 is enough to cover every place in graph $\mathcal{G}$ to make each node learn a good representation.

\begin{figure}[]
\centering
\includegraphics[width=1.0\linewidth]{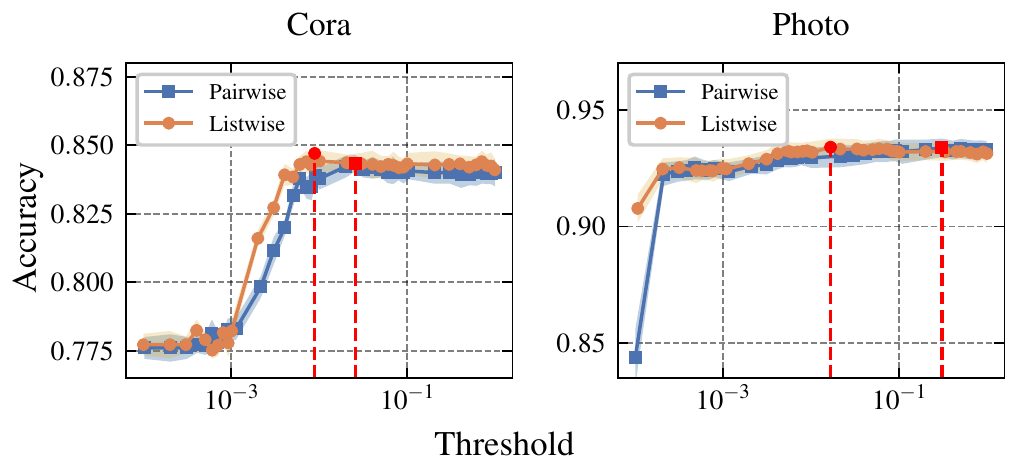}
\caption{Results of GSCL with different thresholds of $\alpha$ and $\beta$ varying in $[0.0001, 1.0]$.}
\label{fig:different_margins}
\end{figure}

\subsection{Study of Neighborhood Sampling Strategies (RQ4)}\label{section: experiment sampling strategy}
To validate the effectiveness of neighborhood sampling strategies, we implement GSCL-Pairwise and GSCL-Listwise with the proposed uniform and weighted sampling strategies, respectively. 
We set the sampling ratio as $20\%$ to allow $20\%$ neighbors in each hop to participate in the calculation of pairwise loss and listwise loss. 
We conduct experiments on WikiCS and Computers as they are representative datasets with the largest average density of neighbors, and the results are reported in Figure~\ref{fig:different_sampling_strategies}.
We can observe that GSCL without any sampling is slightly better than using those two sampling strategies, which is intuitive since sampling results in the loss of information from graphs.
However, GSCL with those two sampling strategies still achieves comparable or even better performance compared with baselines. 
This indicates that our method with neighborhood sampling strategies can be both efficient and effective on large datasets.
Moreover, the uniform sampling strategy always performs better than the Pagerank sampling strategy. 
An potential explanation is that Pagerank tends to select important nodes for the graph, but important nodes in the graph do not necessarily mean that they are more likely to resemble anchor nodes.
For example, in WikiCS dataset, where web pages are nodes, citations are edges, and areas are labels, more significant web pages are more likely to be referenced by pages in a variety of different areas.

\begin{table}[!t]
\centering
\caption{Performance of GSCL adopting different variants  of InfoNCE.}
\setlength{\tabcolsep}{1.5mm}{
\begin{tabular}{@{}lcccc@{}}
\toprule
 \textbf{Method} & \textbf{Cora} & \textbf{Citeseer} & \textbf{WikiCS} & \textbf{Photo} \\ \midrule
\rowcolor{gray!20} GSCL-Pairwise & $84.4$  & $73.7$  &  $80.14$ & $93.42$ \\
\quad \,-InfoNCE$_{\mathrm{out}}$ &  $83.4({\color{red}\downarrow}1.0)$ & $73.1({\color{red}\downarrow}0.6)$ & $78.71({\color{red}\downarrow}1.43)$   & $93.16({\color{red}\downarrow}0.26)$  \\
\rowcolor{gray!20} GSCL-Listwise & $84.5$  & $73.6$  &  $80.16$ & $93.40$ \\
\quad \,-InfoNCE$_{\mathrm{out}}$ & $83.4({\color{red}\downarrow}1.1)$  & $73.3({\color{red}\downarrow}0.3)$ & $78.94({\color{red}\downarrow}1.22)$   & $93.18({\color{red}\downarrow}0.22)$ \\
\bottomrule
\end{tabular}}
\label{table:infonce_var}
\end{table}

\subsection{Study of the InfoNCE Variants of GSCL (RQ5)}
As described in Section~\ref{sec:handling} and Section~\ref{sec:nr}, in order to handle all neighbors from one smaller hop robustly thereby satisfying the overall nature of the diminishing trend described in Proposition \ref{proposition:ranking similarity}, we use InfoNCE$\mathbf{_{in}}$ ({\it i.e.}, Equation~\ref{equation:in}) in a recursive manner to design our GSCL-Pairwise and GSCL-Listwise.
To verify the robustness and effectiveness of such choice, we perform experiments on Cora, Citeseer, WikiCS and Photo datasets to compare GSCL-Pairwise and GSCL-Listwise with their variants which adopt InfoNCE$\mathbf{_{out}}$ ({\it i.e.}, Equation~\ref{equation:out}) to design their corresponding loss functions.
The results are reported in Table~\ref{table:infonce_var}, from which we can observe that both GSCL-Pairwise and GSCL-Listwise perform worse on all 4 datasets (especially in Cora and WikiCS, the decline in results was more than 1\%) when InfoNCE$\mathbf{_{out}}$ is adopted.
This is because of the overall nature of the diminishing trend of label consistency, {\it i.e.}, not all closer/further neighbors will be 100\% more/less similar.
For example, the high-order neighbors could also contain some nodes that lie in the same class of the anchor nodes. 
However, InfoNCE$\mathbf{_{out}}$ will force all closer neighbors to have a higher similarity which may lead to the noisy/false positives.
While InfoNCE$\mathbf{_{in}}$ is not forced to set a higher similarity to all positive pairs and can neglect the noisy/false positives, given that a sufficiently large similarity is set for the true positives~\cite{hoffmann2022ranking,miech2020end}.
Therefore it is more robust and efficient to adopt InfoNCE$\mathbf{_{in}}$ to design the loss functions of GSCL.

\subsection{Study of Threshold $\alpha$ and $\beta$ (RQ6)}\label{section: experiment thresholds}

We also investigate the effects of thresholds $\alpha$ and $\beta$ on GSCL-Pairwise and GSCL-Listwise, respectively. 
We conduct experiments on Cora and Photo and the results are reported in Figure~\ref{fig:different_margins}.
We select the models with the best performance (marked as red in Figure \ref{fig:different_margins}), and fix all other hyperparameters but only vary the thresholds $\alpha$ and $\beta$ in $[0.0001, 1.0]$ to record the accuracy. 
From Figure~\ref{fig:different_margins} we can observe that GSCL performs poorly when the thresholds $\alpha$ and $\beta$ are extremely small, as a too small thresholds limits the maximum similarity between the nodes, resulting in all nodes being less similar.
However, as $\alpha$ and $\beta$ increase, GSCL's ability to control the distance of similarities between neighbors with different hops  diminishes, resulting in a decline in accuracy.
Moreover, the best $\beta$ for GSCL-Listwise is generally smaller than the best $\alpha$ for GSCL-Pairwise. 
The reason is that GSCL-Pairwise performs ranking at a much finer granularity with smaller values in the denominator of the loss function, leading to the need for a larger threshold to control the distance in the embedding space.

\begin{figure}[!t]
\centering
\includegraphics[width=1.0\linewidth]{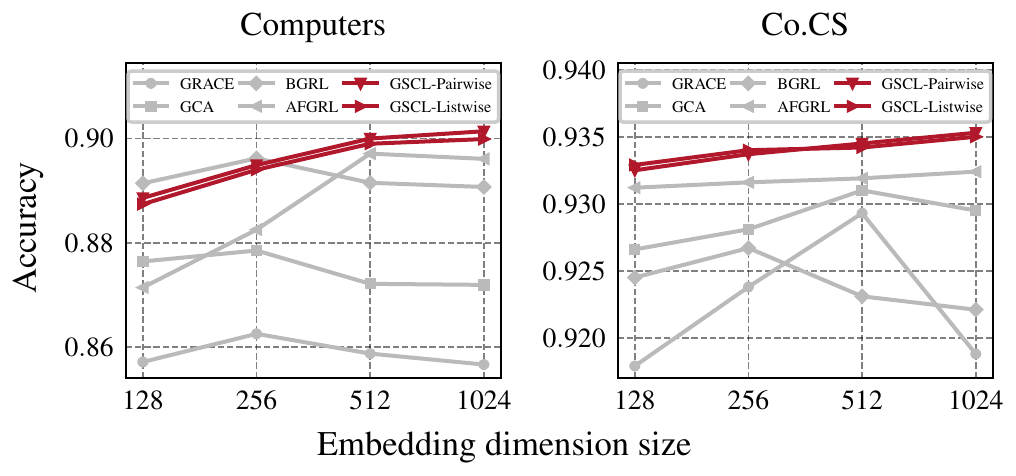}
\vspace{-0.25cm}
\caption{Results of GSCL with different embedding dimension size $D^{'}$ and the comparison with other methods.}
\label{fig:different_dimension}
\end{figure}

\subsection{Effect of Different Embedding Dimension Size (RQ7)}
It is well known that the larger the size of the embedding dimension, the larger the capacity of the model.
Therefore, to evaluate the scalability of GSCL as the model dimension increases, we conduct experiments on Computers and Co.CS across various sizes of node embedding dimension $D^{'}$.
From Figure~\ref{fig:different_dimension} we can observe that both GSCL-Pairwise and GSCL-Listwise are able to benefit more from larger dimensions than other methods.
This indicates that GSCL is more capable of learning especially when compared to GRACE~\cite{Zhu2020DeepGC}, GCA~\cite{Zhu2021GraphCL} and BGRL~\cite{thakoor2021large}, because they saturate their performance with fewer parameters.

\begin{figure*}[!t]
\centering
\includegraphics[width=0.98\linewidth]{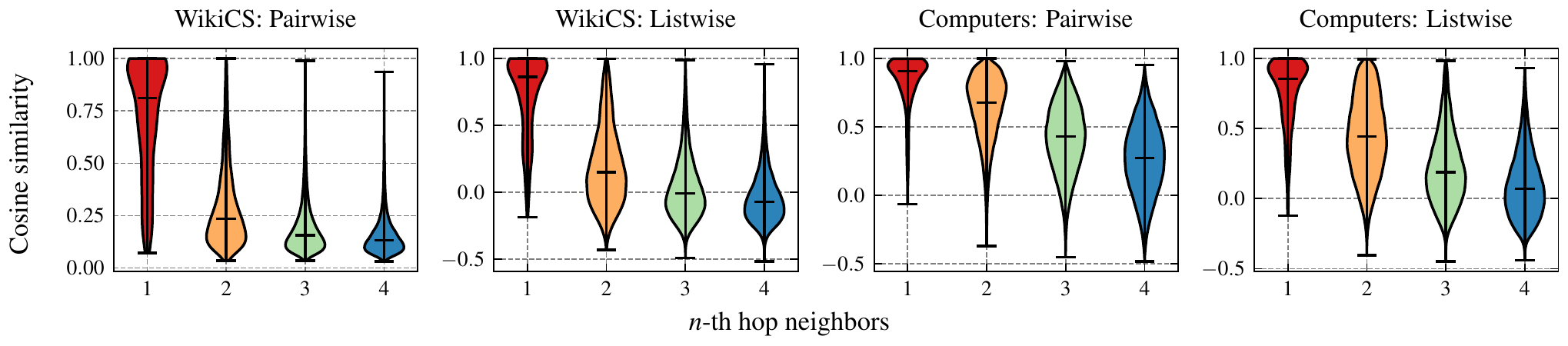}
\vspace{-0.25cm}
\caption{Similarities between the anchor nodes' and associated n-th hop neighbors' representations.}
\vspace{-0.3cm}
\label{fig:similarity_visualization}
\end{figure*}

\subsection{Tests of Whether GSCL Can Fulfill the Proposition \ref{proposition:ranking similarity} (RQ8)}
To further examine whether our proposed GSCL can comply with Proposition \ref{proposition:ranking similarity}, we calculate the cosine similarity between each nodes' and associated $n$-th hop neighbors' representations after training well, and demonstrate the results on WikiCS and Computers as violin plots in Figure~\ref{fig:similarity_visualization}. 
The results show that the similarities overall decreases as the hop count increases, indicating that the learned embeddings of GSCL fulfill the requirement of Proposition \ref{proposition:ranking similarity} that nodes that are closer in the graph are overall more similar than far-distant nodes.
In addition, from Figure~\ref{fig:similarity_visualization} we can observe that a smaller fraction of the closer neighbors are still less similar than the further neighbors, which is consistent with the overall nature of the diminishing trend of label consistency in graph data.

%% file: 6-related-work.tex
\section{Related Work}

\subsection{Graph Representation Learning}
Graph representation learning aims to embed graphs and nodes into low-dimensional embedding spaces while preserving the attributes and structural features of graphs.
Early shallow methods such as DeepWalk \cite{perozzi2014deepwalk}, Node2vec \cite{grover2016node2vec}, and LINE \cite{tang2015line} convert graph data into a sequence of nodes by random walks, and then use language models~\cite{mikolov2013distributed} to learn the node representation.
They overemphasise the proximity of nodes at the expense of the structure of the graph~\cite{Velickovic2019DeepGI}.
G2G~\cite{bojchevski2017deep} adopts a personalized ranking formulation w.r.t. the node distances that exploits the natural ordering of the nodes imposed by the graph structure to learn the embeddings.
The ranking idea of G2G is similar to our proposed GSCL, but the ranking in G2G is to introduce graph structure information in graph Gaussian embedding, while the ranking in GSCL is to design GCL from the graph perspective to make GCL more reasonable.
Moreover, the goal of G2G is to make all closer neighbors must be more similar than all further neighbors, while GSCL describes an overall similarity ranking trend, {\it i.e.},  nodes that are closer in the graph are overall more similar than far-distant nodes.
Therefore, all the motivations, assumptions, methods, and implementations of ranking of G2G and GSCL are different.
Currently, graph neural network (GNN) methods such as GCN~\cite{welling2016semi}, GAT~\cite{Velickovic2018GraphAN}, GIN~\cite{xu2018powerful} have become the de facto standard of graph representation learning, 
they perform message passing iteratively between neighbouring nodes to capture the rich structural information in the graph.
GNN is believed to implicitly assume strong homophily~\cite{zhu2021graph, zhu2020beyond}.

\subsection{Graph Contrastive Learning}
The mainstream graph contrastive learning (GCL) is similar to the contrastive learning in computer vision. 
The first step is to generate two graph views using data augmentation. 
After that, the two graph views are fed into a shared GNN for the purpose of learning representations, which are then optimized using a contrastive learning objective function~\cite{hjelm2018learning,oord2018representation}.
Most GCL use similar graph data augmentation like topology augmentation and feature augmentation~\cite{Velickovic2019DeepGI,Zhu2020DeepGC,thakoor2021large}.
The optimal data augmentation required for different types of graph data is inconsistent, thus making data augmentation an important and sensitive hyper-parameter~\cite{you2020graph,Xia2022SimGRACEAS,Zhu2021AnES}.
To address this issue, some methods use adaptable, dynamic data augmentation for specific graph data~\cite{Zhu2020DeepGC,You2021GraphCL}.
And more recently, some methods have attempted to generate graph views without data augmentation, but instead by adding noise to the hidden embedding space~\cite{Yu2022AreGA,Zhang2022COSTACF} or model parameters~\cite{Xia2022SimGRACEAS}.
Furthermore, {\it SUGRL} feeds the same graph into a multi-layer perceptron and a GCN to generate two positive graph views~\cite{Mo2022SimpleUG}; 
{\it AFGRL} uses the original graph as one view, and generate another view by searching, for each node in the original graph, nodes that can serve as positive samples via k-nearest-neighbor (k-NN) search in the representation space~\cite{Lee2022AugmentationFreeSL}. 
Despite variations in motivations and implementations, the fundamental ideas underlying all these GCL methods are the same, {\it i.e.}, modeling invariance by specifying absolutely similar pairs.
Our proposed GSCL abandons such setting, but performs GCL in a soft and relative manner by preserving the ranking relationship of similarities in the neighborhood.  

%% file: 7-conclusion.tex
\section{Conclusion}
Existing GCL methods always adopt CL idea from computer vision domain: modeling invariance by specifying absolutely similar pairs. 
However, such idea is inappropriate for graph data because of uncontrollable validity of the generated views and the unreliable absolute judgement of similarity between graph views.  
Then, we introduced a novel GCL paradigm GSCL that conducts CL in a soft and relative manner, by ranking node representations in the neighborhood to reflect the overall diminishing trend of label
consistency. 
We propose gated pair-wise and list-wise ranking InfoNCE for implementing GSCL to rank node representations robustly.
We conduct extensive experiments to validate the effectiveness of GSCL. 
The results indicated that GSCL can achieve superior performance compared with 20 state-of-the-art GCL methods on  a variety of graphs and tasks.
We anticipate our contribution would stimulate further research endeavors beyond the traditional GCL paradigm, and we expect more research efforts to exploit more different or better structural properties of graphs like heterogeneity or isomorphism, etc.

%% file: 8-appendix.tex
\section{Detailed Analysis of Complexity}
~\label{appendix: Complexity}

\input{appendix-complexity}

%% file: appendix-complexity.tex
We present this section for supplementing Section~\ref{section:Complexity_Analysis} in details of complexity analysis. 
As discussed, we utilize the calculation times of score function $s(\cdot)$ for measuring the complexity.
Suppose a graph $\mathcal{G}$ includes $N$ nodes, and the average degree is $d$. 
The average $k$-th neighborhood size $|\mathcal{N}(v_i)^{[k]}|=k^d$. 
Next, we introduce how to calculate the complexity of the gated pair-wise and list-wise ranking InfoNCE losses, respectively.

\subsection{Gated Pair-Wise Ranking InfoNCE Loss}
~\label{appendix: pair-wise}
In Equation~\ref{equation:pair-wise ranking infoNCE}, for each anchor node, we will calculate the corresponding ``outer loop'', and each ``outer loop'' includes several ``inner loop'' as:

\begin{flalign}
    \vspace{-4mm}
    \scalemath{0.65}{
        \mathcal{L} = - \sum\limits_{\mathbf{v}_i\in \mathcal{V}} \frac{1}{k} \sum\limits_{j=1}^{k} \underbrace{ \sum\limits_{m=1}^{k-j+1} \log \underbrace{\left[ \min\{\frac{\sum\limits_{\mathbf{h}_\ast \in \mathbb{H}^{[j]}_i}\exp(\frac{s(\mathbf{h}_i, \mathbf{h}_\ast)}{\tau})}{\sum\limits_{\mathbf{h}_\ast \in \mathbb{H}^{[j]}_i}\exp(\frac{s(\mathbf{h}_i, \mathbf{h}_\ast)}{\tau}) + \sum\limits_{\mathbf{h}_\diamond \in \mathbb{H}^{[j+m]}_i}\exp(\frac{s(\mathbf{h}_i, \mathbf{h}_\diamond)}{\tau})}, \alpha \}\right]}_\text{inner loop}}_\text{outer loop}.}
    \label{equation:pair-wise ranking infoNCE analysis}
    \vspace{-4mm}
    \end{flalign}
    
For the $j$-th neighborhood node of the anchor node $v_i$, it includes $k-j+1=k$ rounds of ``inner loop'' for ranking. 
For each ``inner loop'', it includes $d^{j}$ times of $s(\mathbf{h}_i, \mathbf{h}_\ast)$, and the total number for one ``outer loop'' is $d^j+d^{j+1}+\cdots+d^{k+1}$. 
Then, the number of computation of $s(\cdot)$ can be listed as 

\begin{equation}
    \small
    \vspace{-2mm}
    \begin{aligned}
        \overbrace{\qquad\qquad\qquad\qquad\qquad \qquad\qquad\qquad\qquad\qquad\qquad}^{ inner \enspace loop} \\
    \makecell[c]{outer \\loop}\left\{
    \begin{aligned}
        & 1-th\enspace hop \quad d + d^2 \quad d+d^3 \cdots \quad d + d^k \quad d + d^{k+1} \\
        & 2-th\enspace hop \quad d^2+d^3 \quad \cdots \quad d^2 + d^k \quad d^2 + d^{k+1} \\
        & \qquad\vdots \qquad\qquad\qquad \begin{turn}{90}
            $\ddots$
        \end{turn} \\
        & k-th\enspace hop \quad d^k + d^{k+1} \\
    \end{aligned}
    \right.
    \end{aligned}
\label{equation:appendix-complex-pairwise}
\end{equation}

\vspace{2mm}
Then, for the anchor node $v_i$, we need 
\begin{align}
\vspace{-2mm}
\small
\begin{split}
     &k\cdot d + k\cdot d^2 + \cdots + k\cdot d^k + k\cdot + d^{k+1} \\ 
   =& k\cdot(d + d^2 + \cdots + d^k + d^{k+1}) \\
   =& k\sum \limits_{i=1}^{k+1} d^i  .
\end{split}
\end{align}

Since $\mathcal{G}$ contains $N$ nodes, the total computation times of $s(\cdot)$ is $N \cdot k\sum \limits_{i=1}^{k+1} d^i$.
    

\subsection{Gated List-Wise Ranking InfoNCE Loss}
~\label{appendix: list-wise}
In Equation~\ref{equation:list-wise ranking infoNCE}, for each anchor node, we only need to calculate the ``outer loop'', because no ``inner loop'' is in the list-wise ranking.
\begin{equation}
   \small
    \mathcal{L} = - \sum\limits_{\mathbf{v}_i\in \mathcal{V}} \frac{1}{k} \sum\limits_{j=1}^{k} \log \underbrace{\left[ \min \{ \frac{\sum\limits_{\mathbf{h}_\ast \in \mathbb{H}^{[j]}_i}\exp(\frac{s(\mathbf{h}_i, \mathbf{h}_\ast)}{\tau})}{\sum\limits_{j'=j}^{k+1} \sum\limits_{\mathbf{h}_\diamond \in \mathbb{H}^{[j']}_i} \exp(\frac{s(\mathbf{h}_i, \mathbf{h}_{\diamond})}{\tau}) },  \beta \}\right]}_{outer\enspace loop},
    \label{equation:list-wise ranking infoNCE analysis}
\end{equation} 
Therefore, for $j$-th hop neighborhood of the anchor node $v_i$, the ``outer loop'' includes $d^j+d^{j+1}+\cdots+d^{k}+d^{k+1}$ times of $s(\cdot)$. 
Then, the number of computation of $s(\cdot)$ can be listed as 

\begin{equation}
    \small
    \makecell[c]{outer \\loop}\left\{
    \begin{aligned}
        & 1-th\enspace hop \quad d + d^2 + d^3 + \cdots + d^k + d^{k+1}\\
        & 2-th\enspace hop \quad d^2 + d^3 + \cdots + d^k + d^{k+1}\\
        & \qquad\vdots \qquad\qquad\qquad \begin{turn}{90}
            $\ddots$
        \end{turn} \\
        & k-th\enspace hop \quad d^k + d^{k+1}\\
    \end{aligned}
    \right.
    \label{equation:appendix-complex-listwise}
\end{equation} 


Then, for the anchor node $v_i$. we need 

\begin{align}
\small
\begin{split}
    &d+2\cdot d^2 + 3\cdot d^3 + \cdots + k\cdot d^k +d^{k+1} \\
    =& \sum \limits_{i=1}^{k} i\cdot d^i + k\cdot d^{k+1}.
\end{split}   
\end{align}
Since $\mathcal{G}$ contains $N$ nodes, the total computation times of $s(\cdot)$ is $N \cdot (\sum \limits_{i=1}^{k} i\cdot d^i + k\cdot d^{k+1})$.